\documentclass[10pt,twocolumn,letterpaper]{article}

\usepackage{iccv}
\usepackage{times}
\usepackage{epsfig}
\usepackage{graphicx}
\usepackage{amsmath}
\usepackage{amssymb}

\usepackage{subcaption}
\usepackage{pifont}
\usepackage{xcolor}
\usepackage{soul}
\usepackage{epstopdf}
\usepackage[titletoc,title]{appendix}

\usepackage[pagebackref=true,breaklinks=true,letterpaper=true,colorlinks,bookmarks=false]{hyperref}

\iccvfinalcopy 

\def\iccvPaperID{862} 

\ificcvfinal\fi

\newcommand{\cmark}{\ding{51}}%
\newcommand{\xmark}{\ding{56}}%

\DeclareMathOperator{\pool}{pool}
\DeclareMathOperator{\flatten}{flatten}
\DeclareMathOperator{\softmax}{softmax}
\DeclareMathOperator{\reshape}{reshape}
\newcommand{\convfour}{\texttt{conv4}}
\newcommand{\fcseven}{\texttt{fc7}}
\newcommand{\instance}{s}
\newcommand{\im}{I}
\newcommand{\kpclass}{c_{kp}}
\newcommand{\objclass}{c_{o}}
\newcommand{\vp}{\theta}
\newcommand{\gtvp}{\theta_{gt}}
\newcommand{\az}{\theta_1}
\newcommand{\el}{\theta_2}
\newcommand{\ti}{\theta_3}
\newcommand{\anglej}{\theta_j}
\newcommand{\real}{\mathbb{R}}
\newcommand{\imdim}{h \times w \times 3}
\newcommand{\kpmap}{m_{kp}}
\newcommand{\onehotkpclass}{v_{kp}}
\newcommand{\smallkpmap}{m_{pool}}
\newcommand{\smallkpmapvec}{v_m}
\newcommand{\kpmapfeats}{a_m}
\newcommand{\kpmapweights}{W_m}
\newcommand{\kpclassfeats}{a_{\kpclass}}
\newcommand{\kpclassweights}{W_{\kpclass}}
\newcommand{\kpconcatfeats}{a_{kpc}}
\newcommand{\kpconcatweights}{W_{kpc}}

\newcommand{\kpfeats}{a_{kp}}
\newcommand{\convcolumn}{c_{conv4}}
\newcommand{\imkpconcatfeats}{a_{im,kp}}
\newcommand{\imkpconcatweights}{W_{im,kp}}
\newcommand{\angleactivations}{a_{\anglej, \objclass}}
\newcommand{\angleweights}{W_{\anglej, \objclass}}
\newcommand{\instancetuple}{(\im, x, y, \kpclass, \objclass)}
\newcommand{\pthetaofinstance}{P(\vp | s)}
\newcommand{\predrotmatrix}{R_{pr}}
\newcommand{\gtrotmatrix}{R_{gt}}
\newcommand{\rotmatrixdist}{\Delta(\predrotmatrix, \gtrotmatrix)}
\newcommand{\perturbedinstancetuple}{(I, x', y', \kpclass, \objclass)}
\newcommand{\objclassset}{\mathcal{C}_o}
\newcommand{\kpclassset}{\mathcal{C}_{kp} (c_o)}
\newcommand{\acc}{Acc_{\pi/6}}
\newcommand{\accatthresh}{Acc_{Thresh}}
\newcommand{\convweightmap}{\mathcal{W}_\convfour{}}
\newcommand{\hconv}{h_{\convfour}}
\newcommand{\wconv}{w_{\convfour}}
\newcommand{\imfeats}{a_{\fcseven}}
\newcommand{\identity}{\mathbf{I}_2}
\newcommand{\unnormkpmap}{\widehat{\kpmap}}

\begin{document}

\title{Click Here: Human-Localized Keypoints as Guidance for Viewpoint Estimation}

\author{Ryan Szeto and Jason J. Corso\\
Electrical Engineering and Computer Science\\
University of Michigan\\
{\tt\small \{szetor,jjcorso\}@umich.edu}
}

\maketitle
\thispagestyle{empty}

\begin{abstract}
We motivate and address a human-in-the-loop variant of the monocular viewpoint estimation task in which the location and class of one semantic object keypoint is available at test time. In order to leverage the keypoint information, we devise a Convolutional Neural Network called \textnormal{Click-Here CNN (CH-CNN)} that integrates the keypoint information with activations from the layers that process the image. It transforms the keypoint information into a 2D map that can be used to weigh features from certain parts of the image more heavily. The weighted sum of these spatial features is combined with global image features to provide relevant information to the prediction layers. To train our network, we collect a novel dataset of 3D keypoint annotations on thousands of CAD models, and synthetically render millions of images with 2D keypoint information. On test instances from PASCAL 3D+, our model achieves a mean class accuracy of 90.7\%, whereas the state-of-the-art baseline only obtains 85.7\% mean class accuracy, justifying our argument for human-in-the-loop inference.
\end{abstract}

\section{Introduction}

\begin{figure}[t]
	\begin{center}
		\includegraphics[width=\linewidth]{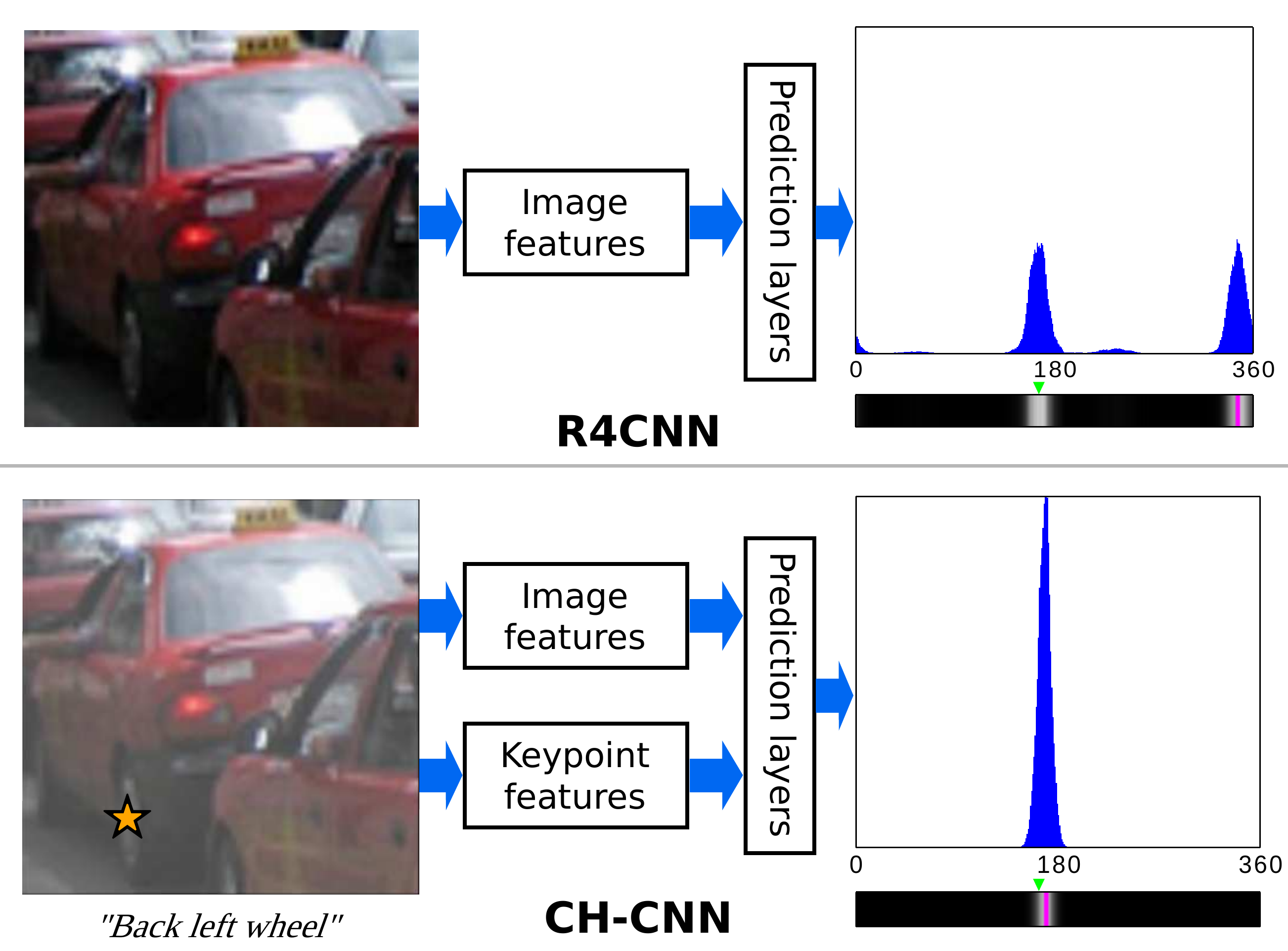}
	\end{center}
    \vspace{-5pt}
	\caption{Semantic keypoint information can help address ambiguities that are difficult to resolve from the image alone. Each diagram shows the available information on the left, the high-level structure of the model in the middle, and the confidences of the azimuth angle on the right. In the black bars, gray indicates confidence, magenta marks the final prediction, and the green triangle marks the ground truth. The orange star indicates the human-provided keypoint. Both the light mask and orange star on the bottom left image are for visualization purposes only, and are not part of the input to any network.}
	\label{fig:keypoint_guidance_motivation}
    \label{fig:pageone}
    \vspace{-10pt}
\end{figure}

It is well understood that humans and computers have complementary abilities.  Humans, for example, are good at visual perception---even in rather challenging scenarios such as finding a toy in a cluttered room---and, consequently, subsequent abstract reasoning from visually acquired information.  On the other hand, computers are good at processing large amounts of data quickly and with great precision, such as predicting viewpoints for millions of images within an exact, but possibly inaccurate, degree.  Although we, as a community, design automatic systems that seek to extract information from images automatically---and have done this quite well, e.g., \cite{he_deep_2016,liu_eccv2016}---there are indeed situations that are beyond the capabilities of current systems, such as inferring the extent of damage to two vehicles involved in a car accident from data acquired by a dash-cam.

In such exceptionally challenging cases, integrating the abilities of both humans and computers during inference is necessary; we call this methodology \textit{hybrid intelligence}, borrowing a term from social computing~\cite{merritt_kurator_2017}. This strategy can lead to pipelines that achieve better performance than fully automatic systems without incurring a significant burden on the human (Figure \ref{fig:pageone} illustrates such an example). Indeed, numerous computer vision researchers have begun to investigate tasks inspired by this methodology, such as learning on a budget \cite{vijayanarasimhan_far-sighted_2010} and Markov Decision Process-based fusion \cite{russakovsky_best_2015}.  

Continuing in this vein of work, we focus on integrating the information provided by a human as additional input during inference to a novel convolutional neural network (CNN) architecture. We refer to this architecture as the \textit{Click-Here Convolutional Neural Network}, or CH-CNN. In training, we learn how to best make use of the additional keypoint information.  We develop a means to encode the location and identity of a single semantic keypoint on an image as the extra human guidance, and automatically learn how to integrate it within the part of the network that processes the image. The human guidance keypoint essentially determines a weighting, or attention mechanism~\cite{xu_show_2015}, to identify particularly discriminative locations of information as data flows through the network. To the best of our knowledge, this is the first work to integrate such human guidance into a CNN at inference time.

To ground this work, we focus on the specific problem of monocular viewpoint estimation---the problem of identifying the camera's position with respect to the target object from a single RGB image. This challenging problem has applications in numerous areas such as automated driving, robotics, and scene understanding, many of which we envision a possible human-in-the-loop during inference. Although discriminative CNN-based methods have achieved remarkable performance on this task~\cite{tulsiani_viewpoints_2015,su_render_2015,li_deep_2016,wu_single_2016}, they often make mistakes when faced with three types of challenges: \textit{occlusion}, \textit{truncation}, and \textit{highly symmetrical objects}~\cite{su_render_2015}. In the first two cases, there is not enough visual information for the model to make the correct prediction, whereas in the third case, the model cannot identify the visual cues necessary to select among multiple plausible viewpoints.

Monocular viewpoint estimation is well-suited to our hybrid intelligence setup as humans can locate semantic keypoints on objects, such as the center of the left-front wheel on a car, fairly easily and with high confidence. CH-CNN is able to integrate such a keypoint directly into the inference pipeline. It computes a distance transform based on the keypoint location, combines it with a one-hot vector that indicates the keypoint class label, and then uses these data to generate a weight map that is combined with hidden activations from the convolutional layers that operate on the image. At a high level, our model learns to extract two types of information---global image information and keypoint-conditional information---and uses them to obtain the final viewpoint prediction.  

We train CH-CNN with over 8,000 computer-aided design (CAD) models from ShapeNet~\cite{chang_shapenet:_2015} annotated with a custom, web-based interface. To our knowledge, our keypoint annotation dataset is an order of magnitude larger than the next largest keypoint dataset for ShapeNet CAD models~\cite{li_deep_2016} in terms of number of annotated models. As our thorough experiments show, we are able to use this human guidance to vastly improve viewpoint estimation performance: on human-guidance instances from the PASCAL 3D+ validation set~\cite{xiang_beyond_2014}, a fine-tuned version of the state-of-the-art model from Su et al.~\cite{su_render_2015} achieves 85.7\% mean class accuracy, while our CH-CNN achieves 90.7\% mean class accuracy. Additionally, our model is well-suited for handling challenges that the state-of-the-art model often fails to overcome, as shown by our qualitative results.

We summarize our contributions as follows. First, we propose a novel CNN that integrates two types of information---an image and information about a single keypoint---to output viewpoint predictions; this model is designed to be incorporated into a hybrid-intelligence viewpoint estimation pipeline. Second, to train our model, we collect keypoint locations on thousands of CAD models, and use these data to render millions of synthetic images with 2D keypoint information. Finally, we evaluate our model on the PASCAL 3D+ viewpoint estimation dataset~\cite{xiang_beyond_2014} and achieve substantially better performance than the leading state-of-the-art, image-only method, validating our hybrid intelligence-based approach. Our code and 3D CAD keypoint annotations are available on our project website at \href{http://ryanszeto.com/projects/ch-cnn}{\texttt{ryanszeto.com/projects/ch-cnn}}.

\begin{figure*}
	\begin{center}
		\includegraphics[width=\linewidth]{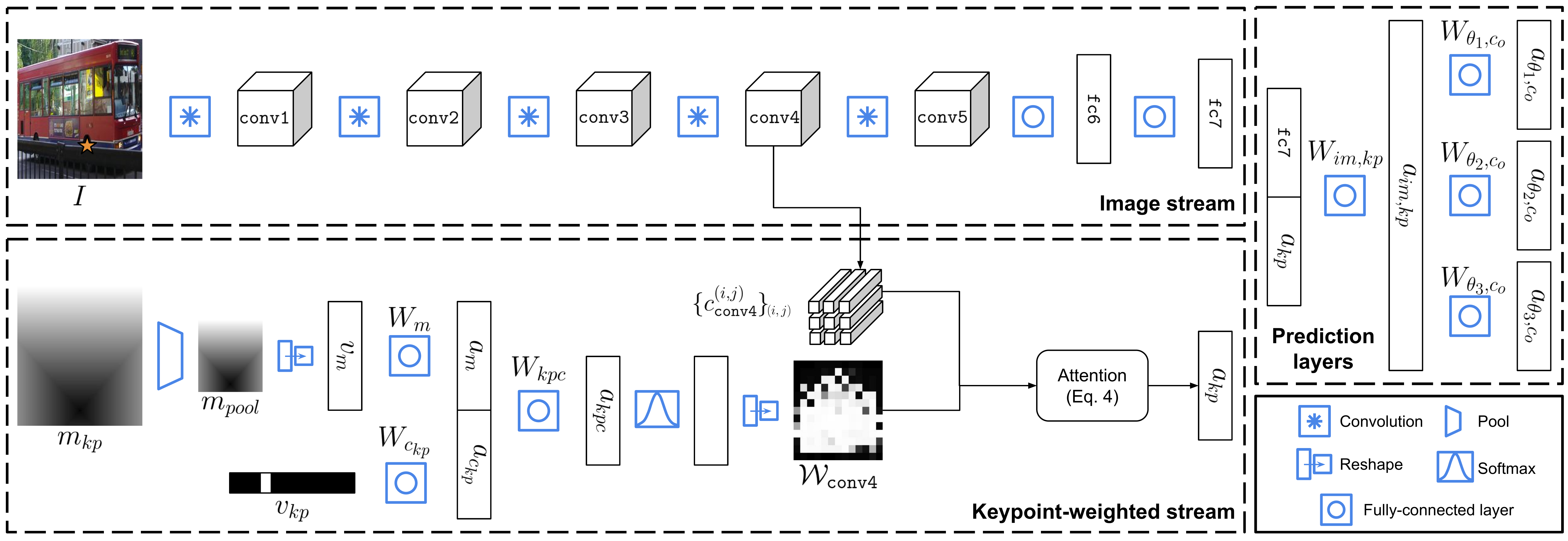}
	\end{center}
	\vspace{-10pt}
	\caption{The architecture for CH-CNN. A weighting over the \convfour{} activation depth columns is learned by taking linear transformations of the keypoint data and applying a softmax operation to the result. The keypoint features are obtained by taking the sum of each activation depth column weighted by the corresponding value in the weight map. These features are concatenated to the \texttt{fc7} image features to aid with inference. The orange star only visualizes the keypoint in this figure; it is not used as input to the network.}
	\vspace{-10pt}
	\label{fig:model_architecture}
\end{figure*}

\section{Related Work}

\noindent \textbf{Monocular Viewpoint Estimation.} Viewpoint estimation and pose estimation of rigid objects have been tackled using a wide variety of approaches. One line of work has extended Deformable Part Models (DPMs)~\cite{felzenszwalb_object_2010} to simultaneously localize objects and predict their viewpoint~\cite{xiang_beyond_2014,pepik_teaching_2012,fidler_3d_2012}. However, DPM-based methods can only predict a limited set of viewpoints, since each viewpoint requires a separate set of models. Patch alignment-based approaches identify discriminative patches from the test image and match them to a database of rendered 3D CAD models~\cite{aubry_seeing_2014,lim_parsing_2013}. More recent approaches have leveraged CNNs~\cite{chen_3d_2015,chen_monocular_2016,wu_single_2016,li_deep_2016,tulsiani_viewpoints_2015,su_render_2015}, which achieve high performance without requiring the hand-crafted features used by earlier work. Additionally, unlike DPM-based approaches, CNNs extend easily to fine-grained viewpoints by regressing from the image to either a continuous viewpoint space~\cite{chen_3d_2015,chen_monocular_2016} or a discrete, but fine-grained space~\cite{tulsiani_viewpoints_2015,su_render_2015}. Even better performance can be achieved by supervising the CNN training stage with intermediate representations~\cite{wu_single_2016,li_deep_2016}. Nonetheless, most fully-automatic approaches struggle from three specific challenges: occlusion~\cite{xiang_beyond_2014,su_render_2015,aubry_seeing_2014}, truncation~\cite{xiang_beyond_2014,su_render_2015}, and highly symmetric objects~\cite{su_render_2015,lim_parsing_2013}. As we show in Section \ref{sec:experiments}, CH-CNN helps reduce the error caused by these challenges.

\noindent \textbf{Human Interaction for Vision Tasks.} Most prior work in the vision community on integrating information from humans at inference time are examples of either active learning or dynamic inference. Active learning approaches reduce the amount of labeled data required for sufficient performance by intelligently selecting unlabeled instances for the human to annotate~\cite{vijayanarasimhan_far-sighted_2010,vondrick_video_2011,vijayanarasimhan_far-sighted_2010,liang_beyond_2014}. Our task differs from active learning in that the information from the human (the keypoint) is available at \textit{inference time} rather than \textit{training time}, and we leverage auxiliary human information to improve the accuracy of our model rather than to achieve sufficient performance with fewer examples. In dynamic inference, a system proposes questions with the goal of improving the confidence or quality of its final answer~\cite{russakovsky_best_2015,branson_visual_2010,wah_multiclass_2011,wah_similarity_2014,jain_click_2016}. This line of work has demonstrated the potential of incorporating human input at inference time. Contrasting with work in dynamic inference, which emphasizes the process of selecting questions for the human to answer, we focus on the problem of learning how to integrate answers in an end-to-end approach for viewpoint estimation CNNs.

\section{Click-Here CNN for Viewpoint Estimation}

Our goal is to estimate three discrete angles that describe the rotation of the camera about a target object, where we are given a tight crop of the object, the location of a visible keypoint in the image, and the keypoint class (e.g. the center of the front right wheel, for a car). We do so with a novel CH-CNN that outputs confidences for each possible angle.

Formally, let $I \in \real^{\imdim}$ be a single RGB image, $(x,y)$ be the 2D coordinate of the provided keypoint location in the image, and $\kpclass$ be the keypoint class. The label $\kpclass$ can take on one of $\sum_{\objclass \in \objclassset} |\kpclassset|$ values, where $\objclassset$ is the set of object classes and $\kpclassset$ is the set of keypoint classes for a given object class $\objclass$. Furthermore, for a given instance $\instance = \instancetuple$, let $\gtvp = (\az, \el, \ti)$ be a tuple associated with $\instance$ representing the ground-truth azimuth/longitudinal rotation, elevation/latitudinal rotation, and in-plane rotation of the camera with respect to the object's canonical coordinate system; each angle is discretized into $N$ bins (following Su et al.~\cite{su_render_2015}, we consider $N=360$). For each object class $\objclass$, we seek a probability distribution function $\pthetaofinstance$ that is maximized at $\gtvp$ for any instance $\instance$. We approximate this set of functions with our CH-CNN.

Prior work~\cite{tulsiani_viewpoints_2015,su_render_2015} has explored the case where $\instance = (\im,\objclass)$, i.e. the image and object class are available at test time, by fine-tuning popular CNN architectures such as AlexNet~\cite{krizhevsky_imagenet_2012} and VGGNet~\cite{simonyan2014very}. Note that after fine-tuning, the intermediate activations of these models can be interpreted as image features that are useful for viewpoint estimation~\cite{su_render_2015}. In our case, we have access to additional information at test time, i.e. the keypoint location $(x,y)$ and class $\kpclass$. We believe that for viewpoint estimation, this information can be used to produce features that complement the global image features extracted from popular CNN architectures. We incorporate this idea in CH-CNN by learning to weigh features from certain regions in the image more heavily based on the keypoint information.

Figure \ref{fig:model_architecture} illustrates the architecture of CH-CNN. The early layers of our architecture are divided into two streams: the first generates features from the image, and the second produces ``keypoint features'' to complement the high-level image features. The keypoint feature stream produces features in three steps. First, a weight map is produced by passing the keypoint map and class through a series of linear transformations and taking the softmax of the result. Second, the activation depth columns from a convolutional layer (\convfour{} in our case) are multiplied by the corresponding weights from the weight map. Finally, the keypoint features are created by taking the sum of the weighted columns.

CH-CNN concatenates the features from the image and keypoint streams and performs inference with one fully-connected hidden layer and one prediction layer for each angle. The fact that we seek a 
probability distribution function for each object class suggests that a separate network must be trained for each object class. To avoid this, we adopt the approach used in Su et al.~\cite{su_render_2015} where lower-level feature layers are shared by all object classes, and object class-dependent prediction layers are used for each angle.

\subsection{Implementation of CH-CNN}
\label{sec:implementation_of_ch-cnns}

We implement the image stream of CH-CNN with the hidden layers of AlexNet~\cite{krizhevsky_imagenet_2012} (i.e. the layers up to the second fully-connected layer \fcseven{}); we take the activations of the \fcseven{} layer as our image features. We stress that while AlexNet is a less powerful model than more recent ones such as ResNet~\cite{he_deep_2016}, our choice allows for a sensible comparison with Su et al.~\cite{su_render_2015}, who fine-tune the same architecture for viewpoint estimation. Additionally, the choice of architecture used for the image stream is independent of our primary contribution, which is to leverage the additional guidance from the provided keypoint at inference time.

The keypoint feature stream takes representations of $(x, y)$ and $\kpclass$ and generates a weighting over activation depth columns from a convolutional layer in the image stream (the fourth layer \convfour{} in our case), where spatial, but high-level information is retained. We use $\convcolumn^{(i, j)}$ to denote the column at position $(i, j)$ in the \convfour{} activation depth column grid. We represent $(x, y)$ with a matrix $\kpmap \in \real^{s \times s}$, where each entry $\kpmap^{(i, j)}$ is the Chebyshev distance of $(i, j)$ from $(x, y)$ divided by the largest possible distance from the keypoint; the label $\kpclass$ is represented with a one-hot vector encoding $\onehotkpclass$.

To learn weights over the activation depth columns, we first learn keypoint map features by downsampling $\kpmap$ with max pooling, and applying a linear transformation to the vectorized result:
\begin{equation}
\begin{split}
    \smallkpmap &= \pool(\kpmap) \\
    \smallkpmapvec &= \flatten(\smallkpmap) \\
    \kpmapfeats &= \kpmapweights \smallkpmapvec
    \enspace.
\end{split}
\end{equation}
Similarly, features from the keypoint class vector are obtained with a linear transformation:
\begin{equation}
    \kpclassfeats = \kpclassweights \onehotkpclass
    \enspace.
\end{equation}
Finally, the weight map for the \convfour{} activation depth columns $\convweightmap$ is obtained by linearly transforming the concatenated keypoint features, applying the softmax function, and reshaping the result to match the shape of the \convfour{} activation depth column grid $(\hconv, \wconv)$:
\begin{align}
    \kpconcatfeats &= \kpconcatweights [\kpmapfeats^\top \, \kpclassfeats^\top]^\top \\
    \convweightmap &= \reshape(\softmax(\kpconcatfeats), (\hconv, \wconv))\nonumber
    \;.
\end{align}
The keypoint feature vector $\kpfeats$ is the sum of the \convfour{} activation depth columns weighted by $\convweightmap$:
\begin{equation}
    \kpfeats = \sum_{i=1}^{\hconv} \sum_{j=1}^{\wconv} \convweightmap^{(i, j)} \convcolumn^{(i, j)}
    \;,
    \label{eq:kpfeats}
\end{equation}
where $i$ and $j$ index into $\convweightmap$ and the \convfour{} activation depth column grid. 

To perform inference, $\imfeats$ and $\kpfeats$ are concatenated. The result is passed through one non-linear hidden layer with an activation function $\sigma$ (e.g. the rectified linear activation function) and a set of class-wise prediction layers for each angle $\anglej$:
\begin{equation}
\begin{aligned}
    \imkpconcatfeats &= \sigma( \imkpconcatweights [\kpfeats^\top \, \imfeats^\top]^\top ) \\
    \angleactivations &= \angleweights \imkpconcatfeats, & j \in \{1, 2, 3\}
    \enspace.
\end{aligned}
\end{equation}

\subsection{Training}
\label{sec:training}

To train our network, we use the geometric structure aware loss function from Su et al.~\cite{su_render_2015},
\begin{align}
	L_{\vp}(S) = -\sum_{s \in S} \sum_{\vp \in \Theta} e^{-d(\vp, \gtvp)/t} \log \pthetaofinstance
    \enspace,
\end{align}
\noindent where $s = \instancetuple$ is a sample from object class $\objclass$, $S$ is the set of training instances, $\Theta$ is the set of possible viewpoints, $\pthetaofinstance$ is the estimated probability of $\vp$ given instance $\instance$, $d(\vp, \gtvp)$ is a distance metric between viewpoints $\vp$ and $\gtvp$ (e.g. the geodesic distance defined in Sec. \ref{sec:comparison_to_image_only_models}), and $t$ is a hyperparameter that tunes the cost of an inaccurate prediction. This loss is a modification of the cross-entropy loss that encourages correlation between the predictions of nearby views.

To train the network, we begin by generating sets of training instances from synthetic data from ShapeNet~\cite{chang_shapenet:_2015} and real-world data from the PASCAL 3D+ dataset~\cite{xiang_beyond_2014} (see Section \ref{sec:data} for details). Then, we initialize the layers from AlexNet with the weights learned from Su et al.~\cite{su_render_2015}; the layers in the keypoint feature stream $\kpmapweights, \kpclassweights, \kpconcatweights$, as well as the prediction layers $\imkpconcatweights$ and $\angleweights$, are initialized with random weights. Next, we train on the synthetic data until the validation performance on a held-out subset of the synthetic data plateaus. Finally, we fine-tune on the real-world training data until the loss on that data plateaus. We develop and train our models in Caffe~\cite{jia_caffe:_2014}. 

\section{Generating Data for CH-CNN}
\label{sec:generating_training_data_for_ch-cnn}
\label{sec:data}

The annotations available in the PASCAL 3D+ dataset~\cite{xiang_beyond_2014} allow us to generate about 14,000 training instances from real-world images (see Section \ref{sec:training_set_details} for details on this process), but this number is insufficient for training CH-CNN. To overcome this limitation, we have extended the synthetic rendering pipeline proposed by Su et al.~\cite{su_render_2015} to generate not only synthetic images with labels, but also 2D keypoint locations, resulting in about two million synthetic training instances. Because this procedure requires knowledge of the 3D keypoint locations on CAD models, we have collected keypoint annotations on 918 bus, 7,377 car, and 320 motorcycle models from the CAD model repository ShapeNet~\cite{chang_shapenet:_2015} with the use of an in-house annotation interface (refer to the supplemental material for details on the CAD model filtering and annotation collection processes). We focus on vehicles to help advance applications in automotive settings, but note that our method is applicable to any rigid object class with semantic keypoints. To the best of our knowledge, the number of annotated CAD models in our dataset is greater than ten times that of the next largest ShapeNet-based keypoint dataset from Li et al.~\cite{li_deep_2016}, who collected keypoints on 472 cars, 80 chairs, and 80 sofas. Our annotated CAD models are publicly available on our project website.

\begin{figure}[t]
	\begin{center}
		\includegraphics[width=0.9\linewidth]{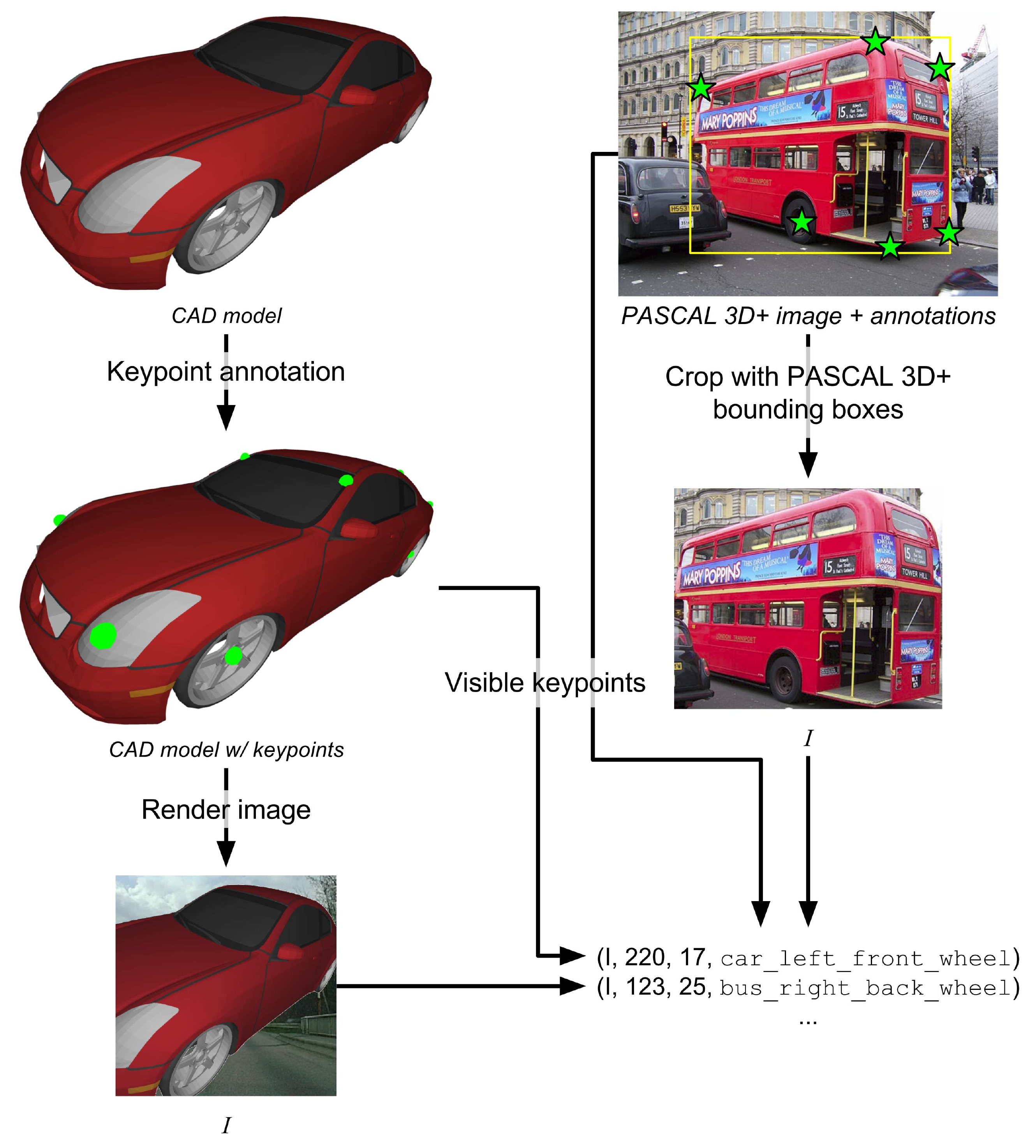}
	\end{center}
	\vspace{-10pt}
	\caption{The pipeline for generating synthetic training data (left) and real-world training data (right).}
	\label{fig:generating_training_data}
	\vspace{-10pt}
\end{figure}

\subsection{Dataset Details}
\label{sec:training_set_details}

We render images of the annotated CAD models using the same pipeline used in Su et al.~\cite{su_render_2015}, which we now describe here. First, we randomly sample light sources and camera extrinsics. Then, we render the CAD model over a random background from the SUN397 dataset~\cite{xiao_sun_2010} to reduce overfitting to synthetic instances. Finally, we crop the object with a randomly perturbed bounding box. From a single rendered image $\im$, we generate one instance of the form $\instancetuple$ with label $\gtvp$ for each visible keypoint, which can be identified by ray-tracing in the rendering environment. We focus on visible keypoints because in the hybrid intelligence environment, we assume that the human locates unambiguous keypoints, which disqualifies occluded and truncated keypoints. We follow this approach to generate about two million synthetic training instances.

PASCAL 3D+ provides detailed annotations that make generating labeled instances a straightforward process. To obtain instance-label pairs from PASCAL 3D+, we extract ground-truth bounding box crops of every vehicle in the dataset. For each cropped vehicle image $\im$ and ground-truth keypoint contained inside $\im$ that is labeled as visible, we produce one labeled instance. We augment the set of training data by horizontally flipping and adjusting $(x, y)$, $\kpclass$, and $\gtvp$ appropriately. In total, we extract about 14,000 training instances and 7,000 test instances from the PASCAL 3D+ training and validation sets, respectively.

\section{Experiments}
\label{sec:experiments}

We conduct experiments to compare image-only viewpoint estimation with our human-in-the-loop approach, as well as analyze the impact of keypoint information on our model. First, we quantitatively compare our model against the state-of-the-art model R4CNN~\cite{su_render_2015} on the three vehicle object classes in PASCAL 3D+ (Section \ref{sec:comparison_to_image_only_models}). Second, we analyze the influence of the keypoint information on our model via ablation tests and perturbations in the keypoint location at inference time (Section \ref{sec:sensitivity_to_keypoint_information}). Finally, we provide qualitative results to compare our model's predictions to those made by R4CNN (Section \ref{sec:qualitative_results}).

\begin{table*}
	\begin{center}
		\resizebox{0.77\linewidth}{!}{
			\begin{tabular}{l | c c c | c || c c c | c}
				\multicolumn{1}{c}{} & \multicolumn{4}{c}{$Acc_{\pi/6}$} & \multicolumn{4}{c}{$MedErr$} \\
				\hline
				& bus & car & motor & \textit{mean} & bus & car & motor & \textit{mean} \\
				\hline
				R4CNN~\cite{su_render_2015} & 92.4 & 78.5 & 81.4 & 84.1 & 5.04 & 7.86 & 14.5 & 9.14 \\
				R4CNN~\cite{su_render_2015}, fine-tuned & 90.6 & 82.4 & 84.1 & 85.7 & 2.93 & 5.63 & 11.7 & 6.74 \\
				Keypoint features (Gaussian fixed attention) & 88.9 & 81.3 & 82.8 & 84.4 & 3.00 & 5.88 & 11.4 & 6.76 \\
				Keypoint features (uniform fixed attention) & 90.6 & 82.0 & 83.7 & 85.4 & 3.01 & 5.72 & 12.1 & 6.93 \\
				\hline
				CH-CNN (keypoint map only) & 90.6 & 82.0 & 84.2 & 85.6 & 3.04 & 5.73 & 11.3 & 6.68 \\
				CH-CNN (keypoint class only) & 90.9 & 86.3 & 83.1 & 86.8 & 2.92 & 5.29 & \textbf{11.0} & 6.41 \\
				CH-CNN (keypoint map + class) & \textbf{96.8} & \textbf{90.2} & \textbf{85.2} & \textbf{90.7} & \textbf{2.64} & \textbf{4.98} & 11.4 & \textbf{6.35} \\
				\hline
			\end{tabular}
		}
	\end{center}
	\vspace{-10pt}
	\caption{PASCAL 3D+ performance for R4CNN~\cite{su_render_2015} with and without fine-tuning on our data, models using a fixed activation depth column weight map, and variants of our CH-CNN model. The CH-CNN models weigh the \convfour{} columns based on the keypoint map, the keypoint class, or both. See Section \ref{sec:comparison_to_image_only_models} for details on the reported metrics.}
	\label{tab:image-only-comparison}
	\vspace{-10pt}
\end{table*}

\subsection{Comparison to Image-Only Models}
\label{sec:comparison_to_image_only_models}

We compare multiple viewpoint estimation models by evaluating their performance on instances extracted from the PASCAL 3D+ validation set~\cite{xiang_beyond_2014}. To be consistent with prior work~\cite{tulsiani_viewpoints_2015,su_render_2015}, we report two metrics, $Acc_{\pi/6}$ and $MedErr$, which are defined as follows. Let $\rotmatrixdist = \frac{||\log(\predrotmatrix^\top \gtrotmatrix)||_F}{\sqrt{2}}$ be the geodesic distance between the predicted rotation matrix $\predrotmatrix$ and the ground-truth rotation matrix $\gtrotmatrix$ on the manifold of rotation matrices. We define $Acc_{\pi/6}$ as the fraction of test instances where $\rotmatrixdist < \pi/6$ in radians, and $MedErr$ as the median value of $\rotmatrixdist$ in degrees over all test instances.

\begin{figure}[t]
	\centering
	\begin{subfigure}[t]{0.23\textwidth}
		\centering
		\includegraphics[width=\linewidth]{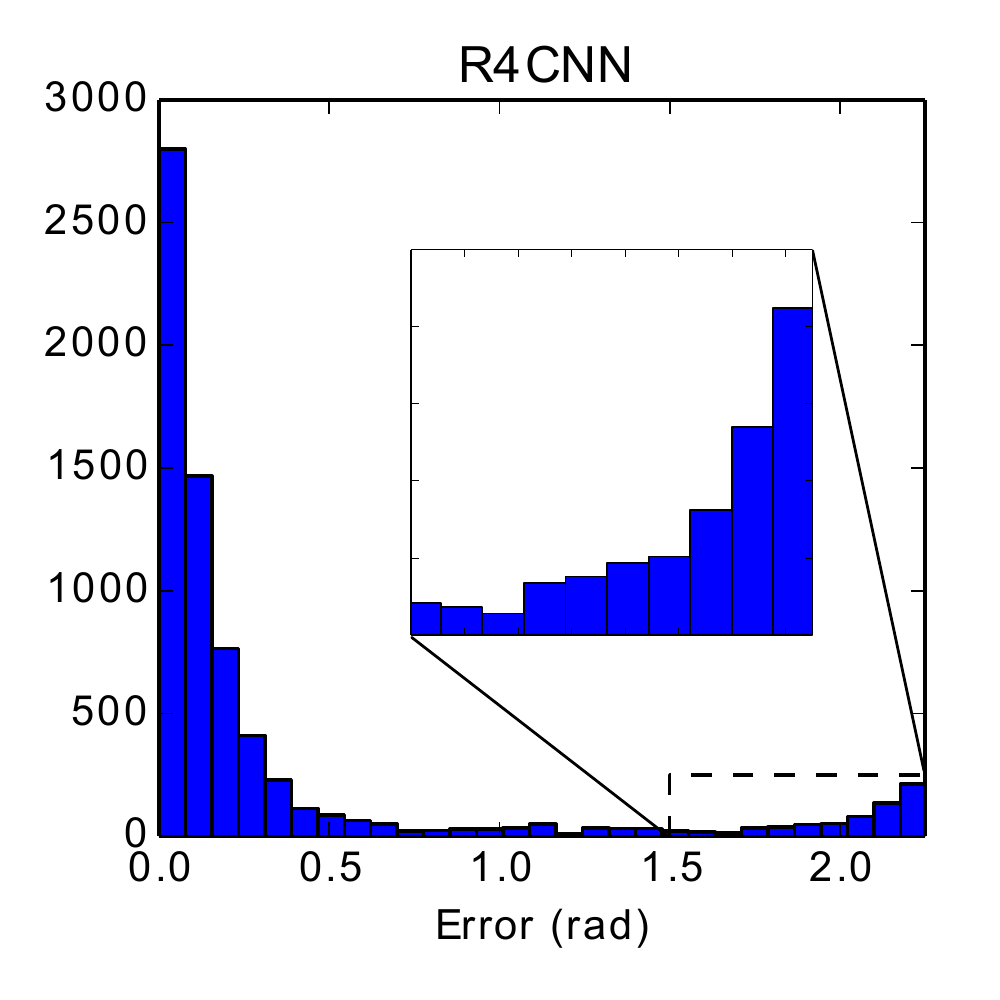}
	\end{subfigure}
	\begin{subfigure}[t]{0.23\textwidth}
		\centering
		\includegraphics[width=\linewidth]{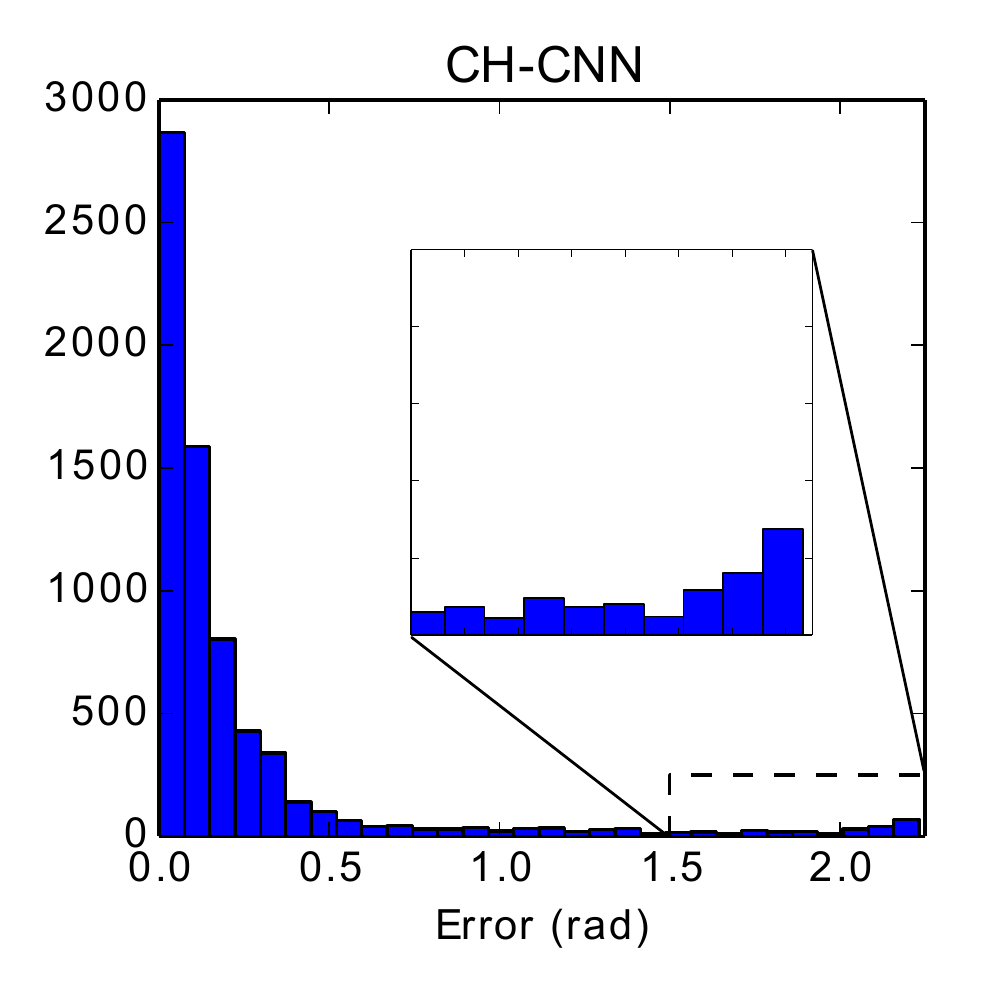}
	\end{subfigure}
	\vspace{-10pt}
	\caption{Distribution of angle error across all classes from fine-tuned R4CNN and our model. In each graph, the area in the dashed box is blown up for clarity.}
	\label{fig:error_distribution}
	\vspace{-15pt}
\end{figure}

Table \ref{tab:image-only-comparison} summarizes the performance of various models on the instances extracted from the PASCAL 3D+ validation set. We include R4CNN with and without fine-tuning (Section \ref{sec:training}) to account for the difference in object classes used in Su et al.~\cite{su_render_2015}. We also compare against two baselines that use a fixed weight map for $\convweightmap$ (Equation \ref{eq:kpfeats}) instead of learning attention from the keypoint data. The first baseline (Gaussian fixed attention) sets $\convweightmap$ to a normalized 13~$\times$~13 Gaussian kernel with a standard deviation of 6, and the second baseline (uniform fixed attention) sets $\convweightmap$ to a 13~$\times$~13 box filter. Aside from the baselines, we evaluate three versions of our CH-CNN model described in Section \ref{sec:implementation_of_ch-cnns}. The first two learn a weight map using either the keypoint map or the keypoint class vector exclusively, and the third is our full model that integrates both sources of information into the weight map computation.

As shown in Table \ref{tab:image-only-comparison}, our full CH-CNN model obtains the highest accuracies out of all tested models by a wide margin; noticeable drops in median error also occur. A conclusion that we draw from these results is that a weighted sum of feature columns can help improve viewpoint estimates. Most importantly, \textit{learning to weigh these features based on the keypoint information is critical to substantially improving performance over image-only methods.} This indicates that providing a single keypoint during inference can indeed help viewpoint estimation by providing features that compliment those extracted solely from the image.

Figure \ref{fig:error_distribution} shows the histograms of angle errors across all object classes obtained by our full CH-CNN model and fine-tuned R4CNN (we refer to this model simply as R4CNN for the remainder of the paper). The most notable difference between the two error distributions occurs along the tails: CH-CNN obtains high errors noticeably less frequently than R4CNN, which we attribute to our model's ability to take advantage of keypoint features when the image features are not informative enough to make a good estimate.

Table \ref{tab:stratified_performance_car_keypoints} stratifies performance by car keypoint classes. In all cases, our model estimates the viewpoint more accurately than R4CNN. However, relative improvement varies greatly, meaning that if certain keypoints can be provided, the improvement from using our model over R4CNN will become more apparent. For instance, CH-CNN yields the greatest relative increase in accuracy when the right back windshield keypoint is provided, but the lowest relative improvement when the right front light keypoint is provided. We attribute this difference to the varying amount of visual information that an image-only system can leverage, which depends on which keypoints are visible: front lights are often more visually distinguishable from their rear counterparts than windshield corners are to their front counterparts. Stratified performance for bus and motorcycle keypoints can be found in the supplementary materials.

\begin{table}
	\begin{center}
		\resizebox{0.85\columnwidth}{!}{
			\begin{tabular}{c | c | c | c }
				\hline
				Keypoint & R4CNN f.t. & CH-CNN & $\%\uparrow$ \\
				\hline
				Left front wheel		& 86.9 & 89.5 & 2.99 \\
				Left back wheel			& 80.6 & 89.0 & 10.4 \\
				Right front wheel		& 89.4 & 91.2 & 2.01 \\
				Right back wheel		& 85.9 & 90.8 & 5.70 \\
				Left front light		& 90.5 & 94.5 & 4.42 \\
				Right front light		& \textbf{93.2} & \textbf{95.5} & \textit{2.47} \\
				Left front windshield	& 87.3 & 91.0 & 4.24 \\
				Right front windshield	& 88.9 & 91.7 & 3.15 \\
				Left back trunk			& 76.8 & 89.5 & 16.5 \\
				Right back trunk		& 72.8 & 88.0 & 20.9 \\
				Left back windshield	& 72.1 & \textit{84.7} & 17.5 \\
				Right back windshield	& \textit{70.8} & 87.6 & \textbf{23.7} \\
				\hline
				\textit{Overall}		& 82.4 & 90.2 & 9.47 \\
				\hline
			\end{tabular}
		}
	\end{center}
	\vspace{-10pt}
	\caption{Values of $\acc$ for the fine-tuned R4CNN model~\cite{su_render_2015} and CH-CNN, stratified by car keypoint class. The $\%\uparrow$ column lists relative percent increase in $\acc$ of CH-CNN over R4CNN. The smallest value in each column is italicized, and the largest value is bolded.}
	\label{tab:stratified_performance_car_keypoints}
	\vspace{-11pt}
\end{table}

\subsection{Sensitivity to Keypoint Information}
\label{sec:sensitivity_to_keypoint_information}

In this section, we explore how changing the keypoint information at inference time affects our trained CH-CNN model. To argue that CH-CNN adapts to the keypoint features rather than ignoring them in favor of the image features, we experiment with providing a keypoint map of all zeros, a keypoint class vector of all zeros, or both to our trained model at test time. As shown in Table \ref{tab:importance-of-keypoint-data}, CH-CNN attains the worst performance when both the keypoint map and class vector are blank. In the cases where either the keypoint map or class is available, but not both, the model achieves better performance. Finally, the best performance is obtained by providing both sources of information. These results indicate that our model adapts to the keypoint information, rather than relying solely on the image features.

\begin{table}
	\begin{center}
		\resizebox{0.9\columnwidth}{!}{
			\begin{tabular}{l | c c | c c c | c}
				\hline
				& KPM & KPC & \textbf{bus} & \textbf{car} & \textbf{mbike} & \textbf{mean} \\
				\hline
				$Acc_{\pi/6}$ & \xmark & \xmark & 75.1 & 67.2 & 80.0 & 74.1 \\
				$Acc_{\pi/6}$ & \xmark & \cmark & 78.0 & 79.4 & 81.8 & 79.7 \\
				$Acc_{\pi/6}$ & \cmark & \xmark & 89.2 & 77.2 & 82.9 & 83.1 \\
				$Acc_{\pi/6}$ & \cmark & \cmark & \textbf{96.8} & \textbf{90.2} & \textbf{85.2} & \textbf{90.7} \\
				\hline
				$MedErr$ & \xmark & \xmark & 3.81 & 8.00 & 12.1 & 7.98 \\
				$MedErr$ & \xmark & \cmark & 3.68 & 6.03 & 12.1 & 7.27 \\
				$MedErr$ & \cmark & \xmark & 2.92 & 6.08 & 11.9 & 6.97 \\
				$MedErr$ & \cmark & \cmark & \textbf{2.64} & \textbf{4.98} & \textbf{11.4} & \textbf{6.35} \\
				\hline
			\end{tabular}
		}
	\end{center}
	\vspace{-10pt}
	\caption{Impact of blank keypoint data on predictions. The KPM and KPC columns respectively indicate whether the ground-truth keypoint map or class was used. \xmark{} indicates that a blank keypoint map or keypoint class vector was used.}
	\vspace{-10pt}
	\label{tab:importance-of-keypoint-data}
\end{table}

\begin{figure}[t]
	\begin{center}
		\includegraphics[width=\linewidth]{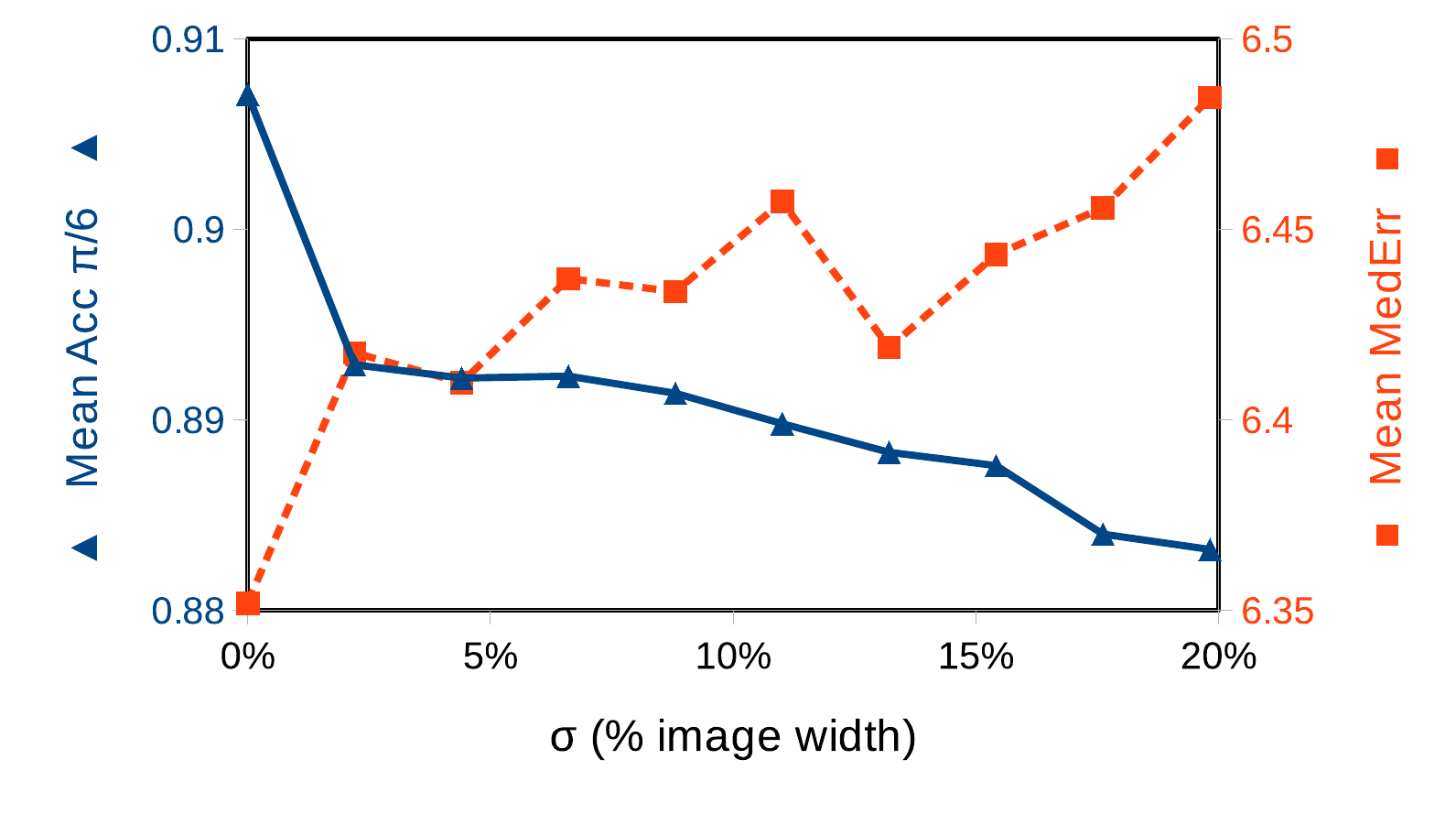}
	\end{center}
	\vspace{-20pt}
	\caption{Sensitivity of CH-CNN to perturbations in the keypoint map. The mean class accuracy is plotted with a solid curve, and the mean class median error is plotted with a dashed curve.}
	\label{fig:keypoint_sensitivity}
	\vspace{-10pt}
\end{figure}

Next, we demonstrate that CH-CNN is robust to noise in the keypoint location at inference time, which is required in order to be useful for the hybrid intelligence environment. The noise is modeled by sampling the keypoint location from a 2D Gaussian whose mean is at the true keypoint location. We accomplish this by creating a new test set for each standard deviation $\sigma$ as follows. We replace each instance $\instancetuple$ from the PASCAL 3D+ validation set with one instance of the form $\perturbedinstancetuple$, where $[x', y']^\top \sim \mathcal{N}([x, y]^\top, \sigma^2 \identity)$. Here, $\identity$ is the 2~$\times$~2 identity matrix and $\sigma$ parameterizes the covariance matrix.

In Figure \ref{fig:keypoint_sensitivity}, we plot the mean class performance of CH-CNN as $\sigma$ increases. We see that our model is robust to misplaced keypoints, retaining over 98\% of its maximum performance even when the standard deviation is about 20\% of the image dimensions. This is likely due to our method of downsampling the keypoint map, which would map the perturbed keypoint to a similar depth column weight map.

\begin{figure}[t]
	\centering
	\begin{subfigure}[t]{\linewidth}
		\centering
		\includegraphics[width=0.4\linewidth]{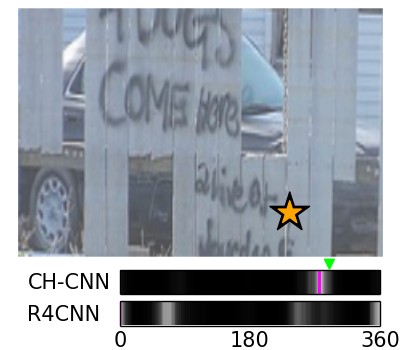}
		\hspace{.3cm}
		\includegraphics[width=0.4\linewidth]{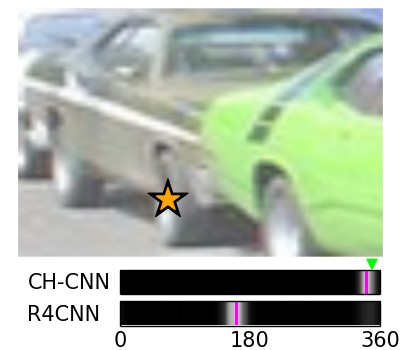}
		\caption{Occlusion}
		\label{fig:qualitative_failure_modes_occlusion}
	\end{subfigure}
	\begin{subfigure}[t]{\linewidth}
		\centering
		\includegraphics[width=0.4\linewidth]{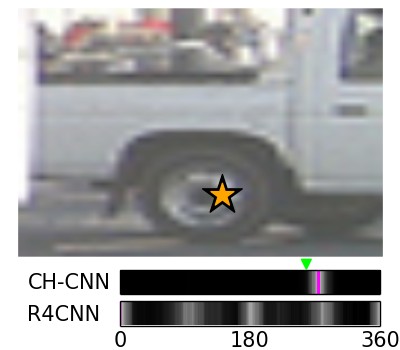}
		\hspace{.3cm}
		\includegraphics[width=0.4\linewidth]{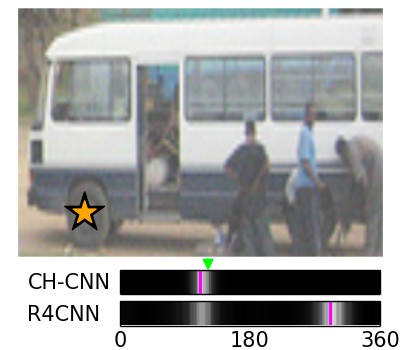}
		\caption{Truncation}
		\label{fig:qualitative_failure_modes_truncation}
	\end{subfigure}
	\begin{subfigure}[t]{\linewidth}
		\centering
		\includegraphics[width=0.4\linewidth]{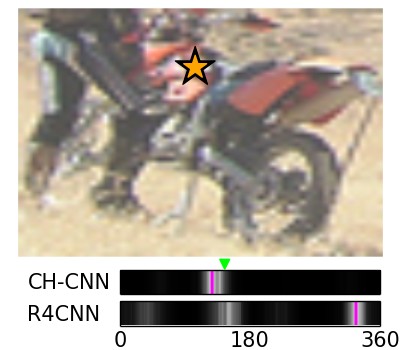}
		\hspace{.3cm}
		\includegraphics[width=0.4\linewidth]{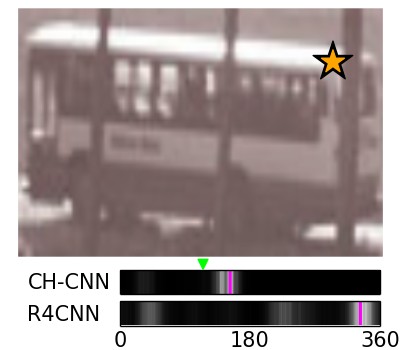}
		\caption{High symmetry}
		\label{fig:qualitative_failure_modes_high_symmetry}
	\end{subfigure}
	\vspace{-6pt}
	\caption{Visualization of challenging instances. Each grayscale bar is the azimuth confidence across all 360 degrees for a model. The green triangle marks the ground truth, and each magenta line marks a final prediction. The light masks and orange stars are for visualizing the keypoint location in this figure only, and are not part of the input to any network.}
	\label{fig:qualitative_failure_modes}
	\vspace{-10pt}
\end{figure}

\subsection{Qualitative Results}
\label{sec:qualitative_results}

\begin{figure*}
	\begin{center}
		\includegraphics[width=.97\linewidth]{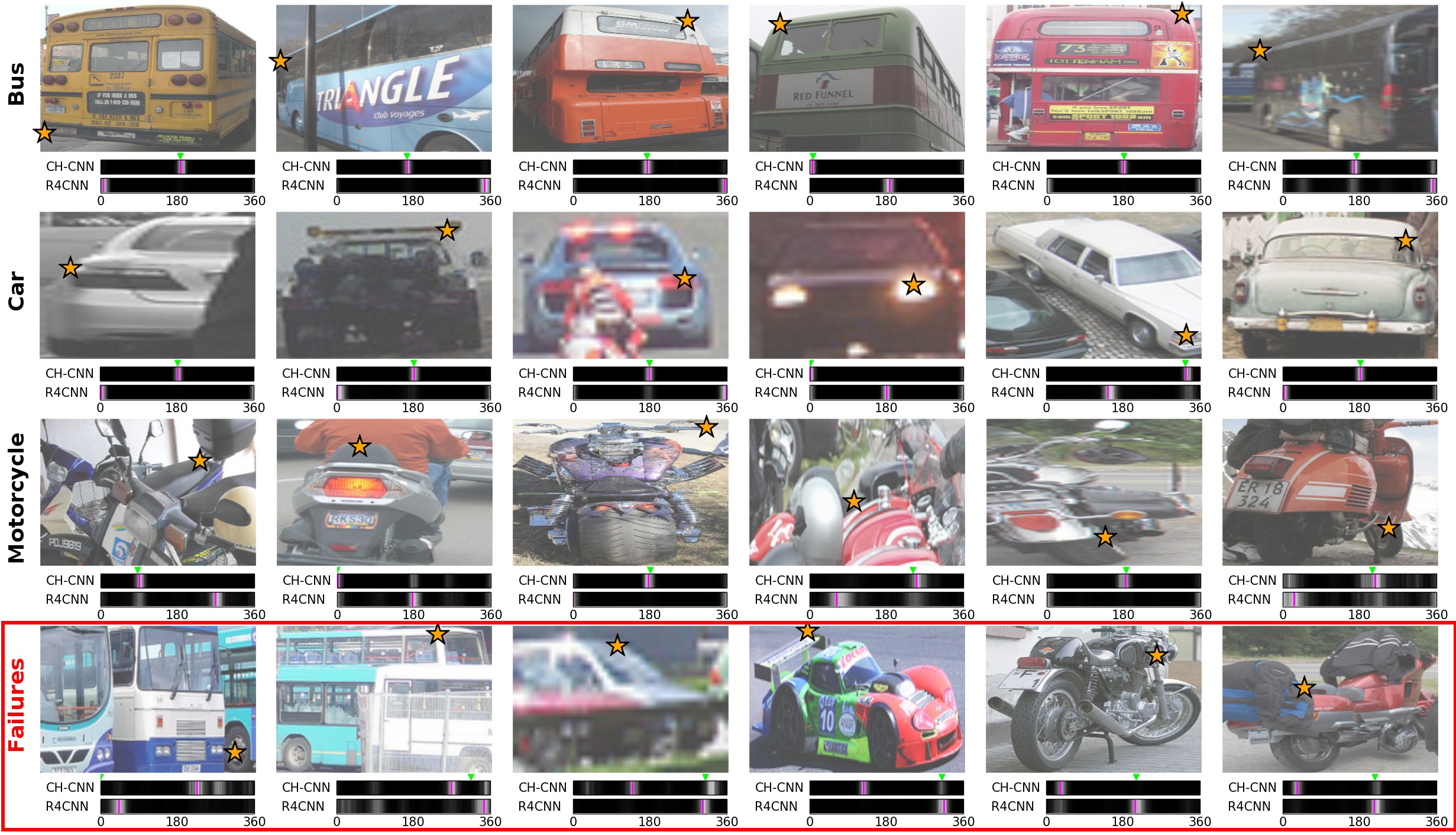}
	\end{center}
	\vspace{-10pt}
	\caption{Azimuth confidences across all object classes, as well as failure cases where our model made an incorrect prediction. See Figure \ref{fig:qualitative_failure_modes} for a description of each plot.}
	\label{fig:visual_comparison}
	\vspace{-9pt}
\end{figure*}

To conclude our analysis, we present qualitative comparisons between CH-CNN and R4CNN~\cite{su_render_2015} by illustrating the confidences across azimuth, the most challenging angle to predict for PASCAL 3D+~\cite{xiang_beyond_2014}. In Figure \ref{fig:qualitative_failure_modes}, we compare the two models for images that exhibit either occlusion, truncation, or highly symmetric objects, observing that CH-CNN tends to estimate viewpoint more robustly than R4CNN under these circumstances. In the shown examples, our model estimates a narrow band around the true azimuth with high confidence. On the other hand, R4CNN exhibits a variety of behaviors, such as multiple peaks (all rows, left), wide bands (middle row, left), or high confidence for the angle opposite the true azimuth (top row, right). We attribute the relative improvement of CH-CNN to the keypoint features, which can help suppress contradictory viewpoint estimates.

Figure \ref{fig:visual_comparison} includes multiple examples of each object class, as well as failure cases for our model. In the positive cases, we continue to see narrower, but more accurate, bands of high confidence from CH-CNN than from R4CNN. Although the negative cases show that CH-CNN does not entirely overcome the main challenges of viewpoint estimation, the improved performance as shown in Table \ref{tab:image-only-comparison} indicates that these factors impact our model less severely than they impact R4CNN.

\section{Conclusion}

\noindent \textbf{Limitations and Suggestions.} Our work makes a few critical assumptions that are worth addressing in future work. First, we assume that information about only one keypoint is provided; in reality, we should be able to leverage multiple keypoints to further improve the estimate. Second, we assume that viewpoint estimates of the same object with different keypoint data are unrelated, whereas a better approach would be to enforce the consistency of viewpoint estimates of the same object. Third, we assume that the provided keypoint is both unoccluded and within the object bounding box. However, this is sensible in the context of hybrid intelligence because we can trust the human to suggest unambiguous keypoints or indicate that none exist, in which case we can fall back on image-only systems.

\noindent \textbf{Summary.} We have presented a hybrid intelligence approach to monocular viewpoint estimation called CH-CNN, which leverages keypoint information provided by humans at inference time to more accurately estimate the viewpoint. Our method combines global image features with keypoint-conditional features by learning to weigh feature activation depth columns based on the keypoint information. We train this model by generating synthetic examples from a new, large-scale 3D keypoint dataset. As shown by our experiments, our method vastly improves viewpoint estimation performance over state-of-the-art, image-only systems, validating our argument that applying hybrid intelligence to the domain of viewpoint estimation can yield great benefits with minimal human effort. To spur further work in hybrid intelligence for 3D scene understanding, we have made our code and keypoint annotations available at \href{http://ryanszeto.com/projects/ch-cnn}{\texttt{ryanszeto.com/projects/ch-cnn}}.

\noindent \textbf{Acknowledgements.} We thank Vikas Dhiman, Luowei Zhou, and Madan Ravi Ganesh for their helpful discussions and management of computing resources. We also thank Alex Miller, Matthew Dorow, Bhavika Reddy Jalli, Hojun Son, Guangyu Wang, Ronald Scott, and the other student annotators for collecting the keypoint dataset. This work was partially supported by the Denso Corporation, NSF CNS 1463102, and DARPA W31P4Q-16-C-0091.

{\small
	\bibliographystyle{ieee}
	\bibliography{egbib}
}

\begin{appendices}
	\onecolumn

\section{Introduction}

This document constitutes the written portion of the supplementary material for \textit{Click Here: Human-Localized Keypoints as Guidance for Viewpoint Estimation}. It is organized as follows:

\begin{itemize}
	\item Appendix \ref{sec:ch-cnn_architecture_and_training_details} provides details for our CH-CNN architecture, including layer sizes and training parameters.
	\item Appendix \ref{sec:keypoint_annotation_collection_details} describes how we collected and verified the CAD model keypoint annotations in our dataset.
	\item Appendix \ref{sec:additional_quantitative_results} provides additional quantitative analysis. We analyze the $\acc$ and $MedErr$ evaluation metrics on our model and R4CNN~\cite{su_render_2015} in multiple ways, such as comparing performance by keypoint class and comparing accuracy over a range of thresholds. Additionally, we list the evaluation metrics on variations of our model that use different types of keypoint maps.
	\item Appendix \ref{sec:additional_qualitative_results} provides additional qualitative results. First, we visualize challenging instances in which occlusion, truncation, and/or symmetry occur. Then, we visualize the weight maps produced by instances from each object class.
\end{itemize}

\section{CH-CNN Architecture and Training Details}
\label{sec:ch-cnn_architecture_and_training_details}

CH-CNN takes three inputs that are generated from instance tuple $\instancetuple$: $\im$, $\kpmap$, and $\onehotkpclass$. $\im$ is a 227 $\times$ 227 $\times$ 3 RGB image subtracted by the ImageNet image mean~\cite{deng2009imagenet}; $\kpmap$ is a 227 $\times$ 227 grayscale image whose values are produced by any method described in Appendix \ref{sec:additional_quantitative_results}; and $\onehotkpclass$ is a 34-length vector (corresponding to 12 bus, 12 car, and 10 motorcycle keypoint classes) with value 1 at keypoint class index $\kpclass$ and zero elsewhere. The image stream of CH-CNN is implemented with the first seven layers from AlexNet~\cite{krizhevsky_imagenet_2012} using the reference architecture available in Caffe~\cite{jia_caffe:_2014}.

To obtain $\kpmapfeats$, the keypoint map is first downsampled with a max pooling layer with a stride and kernel size of 5. Then, the result is flattened and multiplied by the learned 2116 $\times$ 2116 matrix $\kpmapweights$. To obtain $\kpclassfeats$, the one-hot keypoint class vector $\onehotkpclass$ is multiplied by the learned 34 $\times$ 34 matrix $\kpclassweights$. The concatenated vector $[\kpmapfeats^\top \kpclassfeats ^\top]^\top$ is multiplied by the learned 169 $\times$ 2150 weight matrix $\kpconcatweights$ to get $\kpconcatfeats$, to which a softmax and reshaping is applied to obtain a 13 $\times$ 13 weight map whose entries sum to 1 (13 $\times$ 13 comes from the height and width of the \texttt{conv4} activation tensor, and 169 comes from their product). For the inference layers, the hidden activations $\imkpconcatfeats$ are obtained with a learned non-linear fully-connected layer with 4096 outputs and ReLU as the non-linear activation function. Finally, the angle prediction layer for each angle $\anglej$ and object class $\objclass$ takes $\imkpconcatfeats$ and multiplies it by the learned 360 $\times$ 4096 matrix $\angleweights$.

The entire CH-CNN architecture is trained end-to-end with the Adam algorithm~\cite{kingma2014adam} while training on synthetic and real instances. In both cases, the batch size, base learning rate, first momentum rate, and second momentum rate are set to 192, $10^{-4}$, 0.9, and 0.999 respectively. It takes about 3 days on an NVIDIA Titan X Pascal GPU to train CH-CNN.

\section{Keypoint Annotation Collection Details}
\label{sec:keypoint_annotation_collection_details}

\begin{figure}[t]
	\centering
	\begin{subfigure}[t]{.48\linewidth}
		\centering
		\includegraphics[width=\linewidth]{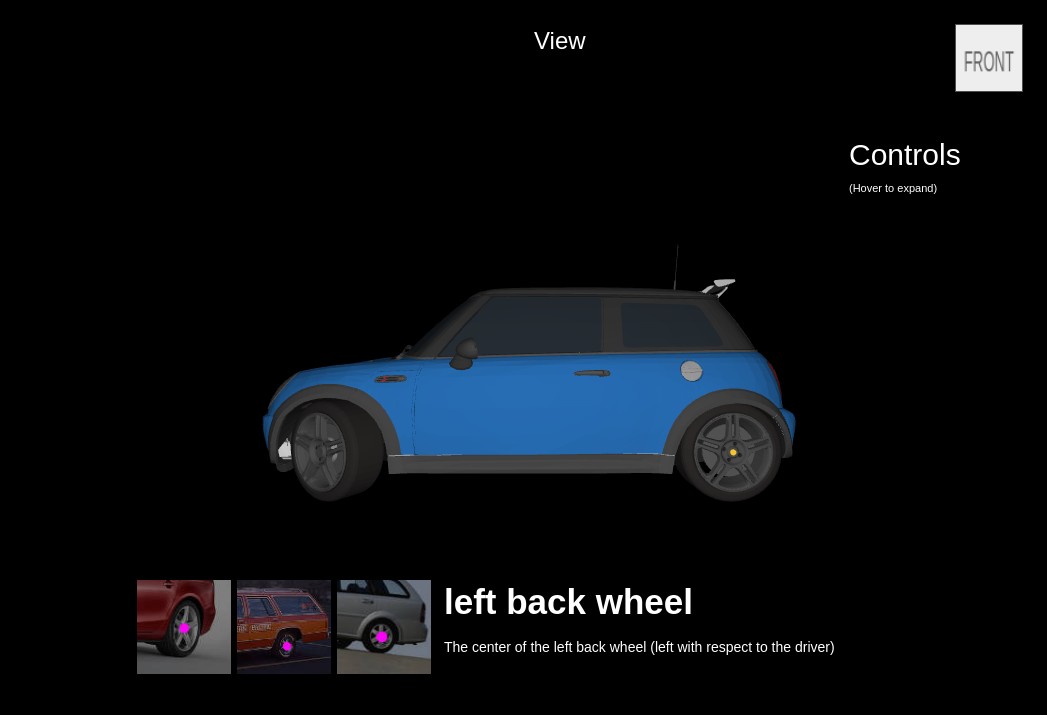}
		\caption{Edit mode}
	\end{subfigure}
	\hfill
	\begin{subfigure}[t]{.48\linewidth}
		\centering
		\includegraphics[width=\linewidth]{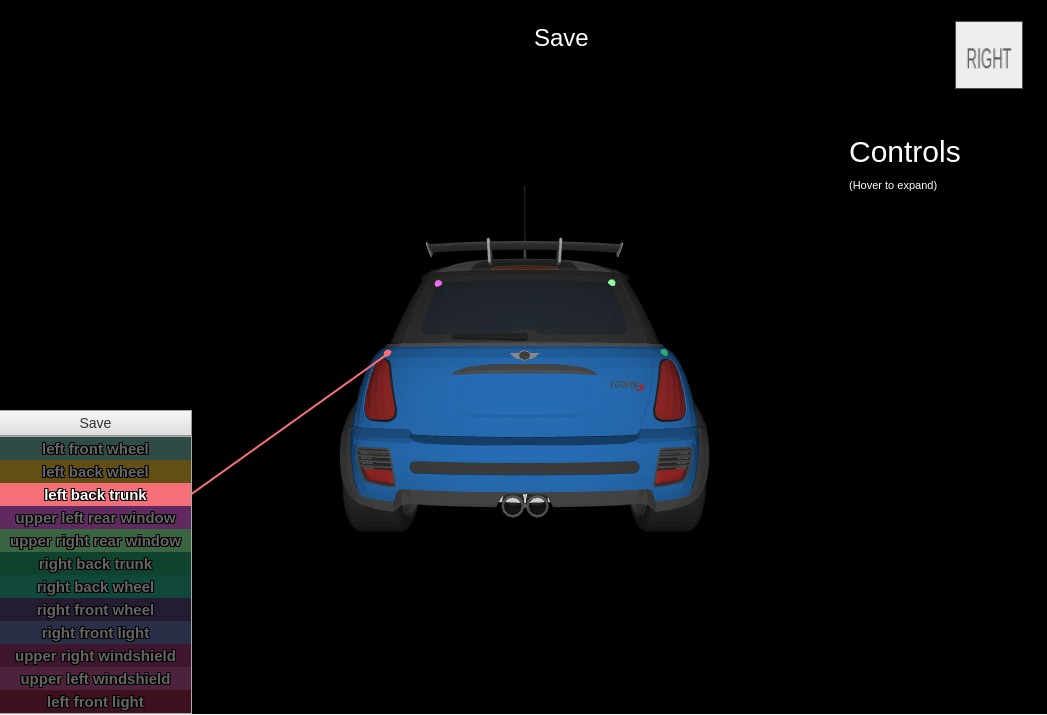}
		\caption{Save mode}
	\end{subfigure}
	\caption{Screenshots of our CAD model annotation interface. Figure best viewed zoomed-in on a monitor.}
	\label{fig:keypoint_annotation_interface}
	\vspace{-10pt}
\end{figure}

Collecting a large number of keypoint annotations efficiently requires a scalable and easily-accessible interface. To this end, we extended the open-source project \textit{cad.js}\footnote{\url{https://github.com/ghemingway/cad.js}}, a web-based interface and server for viewing 3D CAD models, to support keypoint annotation. Figure \ref{fig:keypoint_annotation_interface} shows screenshots of our keypoint annotation interface. When a CAD model is loaded, the user can navigate around the object via rotate, pan, and zoom operations with the mouse. At the bottom of the screen is a panel that describes the requested keypoint with visual examples and text. In order to label the keypoint, the user enters \textit{edit mode} and drags a small sphere onto the appropriate location. The user cycles through and labels all keypoints for the model's object class, then enters \textit{save mode} to preview the annotations before saving them to the server. The preview displays all keypoints at once, with a line drawn from the textual label to the keypoint location; users can view individual keypoints by mousing over the textual label.

To obtain the 3D CAD models, we downloaded the bus, car, and motorcycle models from ShapeNet~\cite{chang_shapenet:_2015}. We restricted ourselves to these three vehicle classes from PASCAL 3D+ due to cost constraints; however, we note that our methods extend naturally to any number of object classes. After filtering out models that did not capture realistic appearance (e.g. models without wheels, cars without bodies, etc.), which comprised about 1.8\% of ShapeNet vehicle models, we were left with 918 bus, 7,377 car, and 320 motorcycle models to annotate.

We hired 10 student annotators over the course of one month to label the 3D CAD models with the semantic keypoints identified in the PASCAL 3D+ dataset~\cite{xiang_beyond_2014}. Although Xiang et al.~\cite{xiang_beyond_2014} identified keypoints meant to broadly describe the corresponding object classes, we found that not all CAD models contained all semantic keypoints (e.g. convertible cars do not have rear windows, so the ``rear windshield corner'' keypoints have no meaning for these models). In these cases, the annotators were instructed to not label those keypoints; as a result, some CAD models do not have an annotation for every keypoint for their object class.

To improve the quality and consistency of annotations, each CAD model was viewed by one annotator and one verifier---the annotator placed the labels, and the verifier checked that the labels were placed appropriately. The verifier sent the model back to the annotator if he/she disagreed with the locations; if they could not reach a mutual agreement on keypoint locations, we annotated the model ourselves.

\section{Additional Quantitative Results}
\label{sec:additional_quantitative_results}

We begin by exploring the impact of using four different types of keypoint maps $\kpmap$ on viewpoint estimation performance for CH-CNN. We produce unnormalized keypoint maps $\unnormkpmap$ defined by the following four procedures:

\begin{itemize}
	\item \textbf{Gaussian keypoint map.} Given keypoint location $(x, y)$, the unnormalized keypoint map $\unnormkpmap$ is given by a 2D Gaussian whose mean is $[x, y]^\top$ and whose standard deviation is about 10\% of the image (23 pixels for our 227 $\times$ 227 images).
	\item \textbf{Euclidean distance transform keypoint map.} Given keypoint location $(x, y)$, each entry in the keypoint map $\kpmap^{(i, j)}$ is given as $$\kpmap^{(i, j)} = \sqrt{(i-x)^2 + (j-y)^2} \enspace.$$
	\item \textbf{Manhattan distance transform keypoint map.} Given keypoint location $(x, y)$, each entry in the keypoint map $\kpmap^{(i, j)}$ is given as $$\kpmap^{(i, j)} = |i-x| + |j-y| \enspace.$$
	\item \textbf{Chebyshev distance transform keypoint map.} Given keypoint location $(x, y)$, each entry in the keypoint map $\kpmap^{(i, j)}$ is given as $$\kpmap^{(i, j)} = \max(|i-x|, |j-y|) \enspace.$$
\end{itemize}

\noindent We produce the final keypoint map $\kpmap$ from an unnormalized keypoint map $\unnormkpmap$ by dividing $\unnormkpmap$ by the maximum possible value over all possible $\unnormkpmap$ (note that this is not necessarily the maximum value of the given $\unnormkpmap$). The performance of CH-CNN with each type of keypoint map is shown in Table \ref{tab:keypoint_map_comparison}.

Table \ref{tab:stratified_performance} lists the performance of fine-tuned R4CNN~\cite{su_render_2015} and CH-CNN on each keypoint type, as well as the relative improvement of our model over fine-tuned R4CNN. From the overall relative improvement for all three object classes under both evaluation metrics, we observe that providing keypoint information generally increases viewpoint estimation performance over R4CNN. However, we also note that in the motorcycle class, some keypoint classes appear to confuse CH-CNN and yield a relative decrease in performance.

In Table \ref{tab:stratified_error}, we compare the errors made by CH-CNN and R4CNN based on per-instance performance rather than aggregate performance. To do this, we compute the error $\rotmatrixdist$ for one instance from R4CNN, and subtract the corresponding value from CH-CNN. The values in Table \ref{tab:stratified_error} are the means of the resulting difference in errors stratified by keypoint class. From this table, we see that across most keypoint classes, CH-CNN predicts an angle closer to the ground truth than R4CNN for any particular instance. We see the same general trends as those seen in Table \ref{tab:stratified_performance}, such as performance varying depending on keypoint class and decreased performance for certain motorcycle keypoint classes.

\begin{figure}[t]
	\begin{center}
		\includegraphics[width=0.7\linewidth]{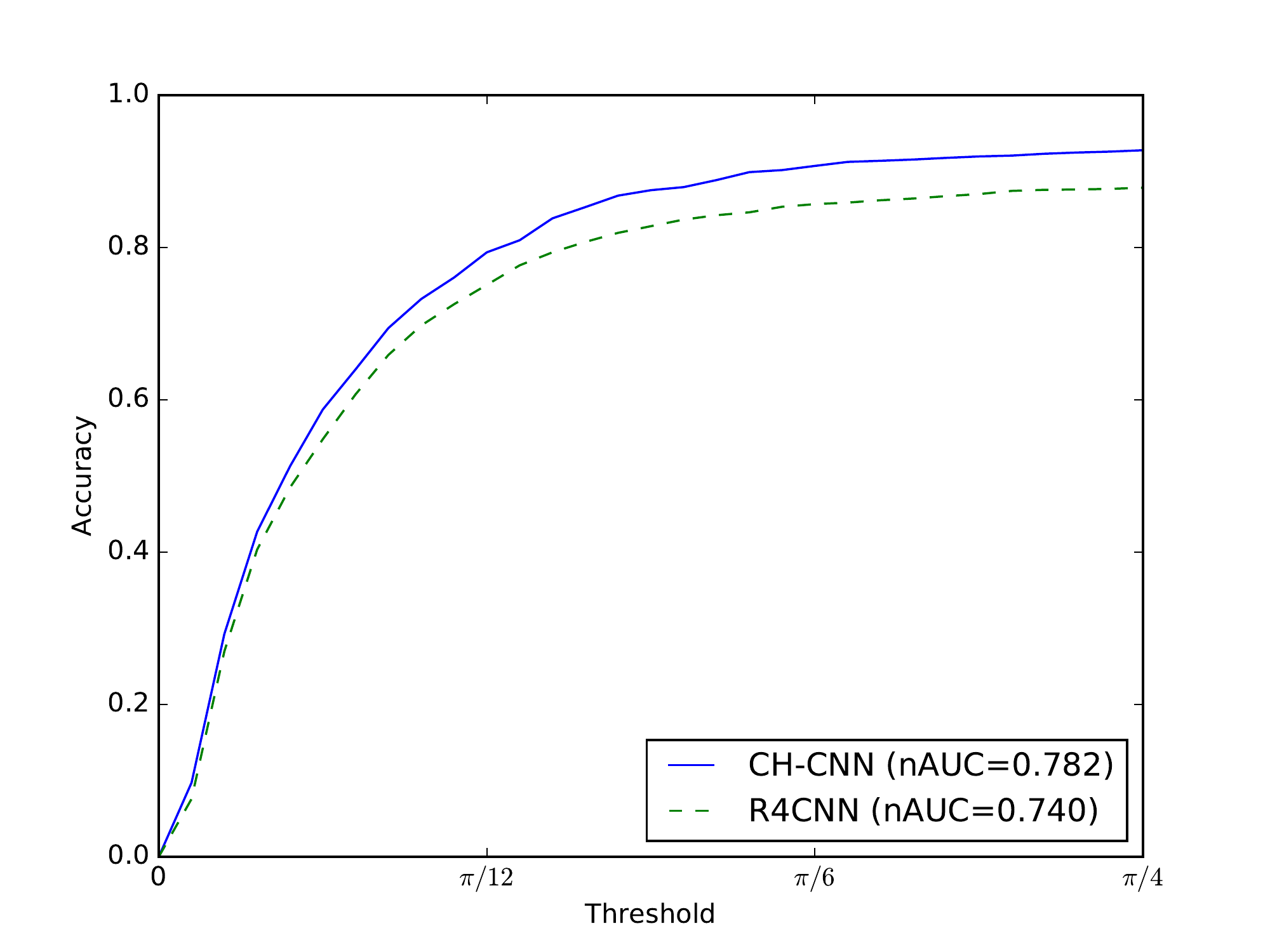}
	\end{center}
	\vspace{-10pt}
	\caption{The metric $\accatthresh$ of R4CNN and CH-CNN for values of $Thresh$ from 0 to $\pi/4$. For a given curve, nAUC is the area under the curve divided by the total plotted area ($\pi/4$ in this graph).}
	\label{fig:acc_at_X}
	\vspace{-10pt}
\end{figure}

In Figure \ref{fig:acc_at_X}, we plot the value of $\accatthresh$, defined as the fraction of test instances where $\rotmatrixdist < Thresh$ in radians, across multiple values of $Thresh$ between 0 and $\pi/4$. We also report the normalized area under the curve (nAUC), which is the percentage of the plotted area that falls under a given curve. From this graph, we notice that the gap in performance between CH-CNN and R4CNN widens considerably with a large enough threshold. However, performance between the two models is similar at very small values of $Thresh$, which suggests the need to focus on improvements at strict threshold values.

\begin{table}[t]
	\begin{center}
		\begin{tabular}{l | c c c | c || c c c | c}
			\multicolumn{1}{c}{} & \multicolumn{4}{c}{$Acc_{\pi/6}$} & \multicolumn{4}{c}{$MedErr$} \\
			\hline
			& bus & car & motor & \textit{mean} & bus & car & motor & \textit{mean} \\
			\hline
			CH-CNN (Gaussian) & 88.9 & 82.2 & 82.1 & 84.4 & 2.92 & 6.05 & 11.5 & 6.81 \\
			CH-CNN (Euclidean) & 94.5 & 89.6 & 84.9 & 89.7 & 2.93 & 5.00 & \textbf{11.0} & \textbf{6.31} \\
			CH-CNN (Manhattan) & 95.4 & \textbf{90.8} & 83.1 & 89.8 & 2.99 & \textbf{4.97} & 11.5 & 6.50 \\
			CH-CNN (Chebyshev) & \textbf{96.8} & 90.2 & \textbf{85.2} & \textbf{90.7} & \textbf{2.64} & 4.98 & 11.4 & 6.35 \\
			\hline
		\end{tabular}
	\end{center}
	\vspace{-10pt}
	\caption{PASCAL 3D+ performance for our CH-CNN model with different keypoint maps. \textit{CH-CNN (Gaussian)} refers to our model using a Gaussian centered around the keypoint location as the keypoint map. The remaining three rows correspond to our model using distance transform keypoint maps with Euclidean, Manhattan, and Chebyshev distance respectively. The best performance in each column (highest value for $\acc$, lowest value for $MedErr$) is bolded.}
	\label{tab:keypoint_map_comparison}
\end{table}

\section{Additional Qualitative Results}
\label{sec:additional_qualitative_results}

\begin{figure}[t]
	\centering
	\begin{subfigure}[t]{\linewidth}
		\includegraphics[width=.24\linewidth]{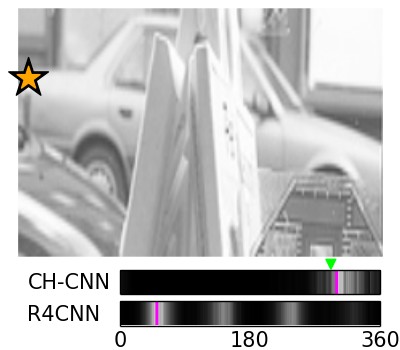}
		\includegraphics[width=.24\linewidth]{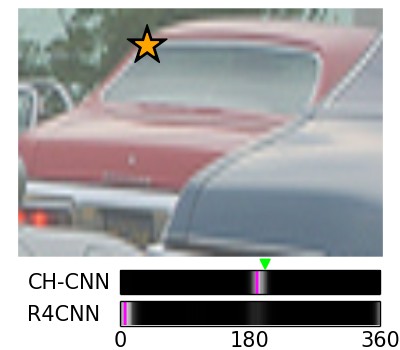}
		\includegraphics[width=.24\linewidth]{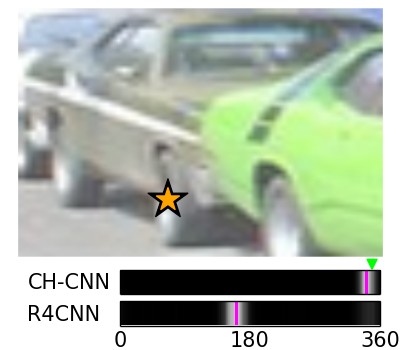}
		\includegraphics[width=.24\linewidth]{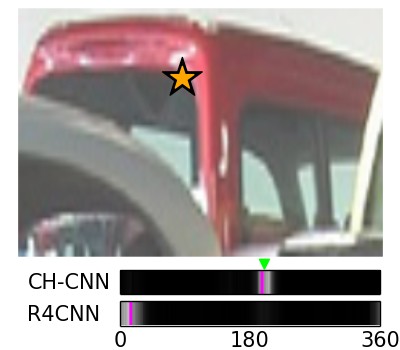}
		\caption{Occlusion}
		\label{fig:challenging_examples_occlusion}
	\end{subfigure}
	\begin{subfigure}[t]{\linewidth}
		\includegraphics[width=.24\linewidth]{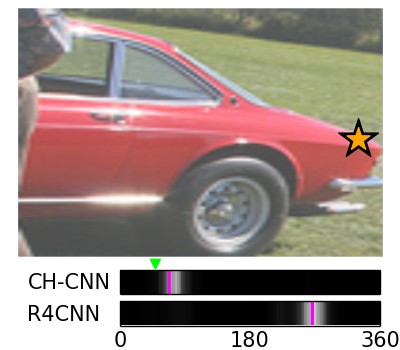}
		\includegraphics[width=.24\linewidth]{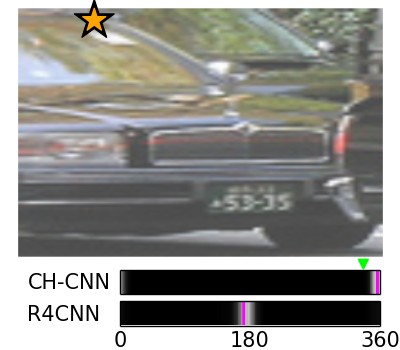}
		\includegraphics[width=.24\linewidth]{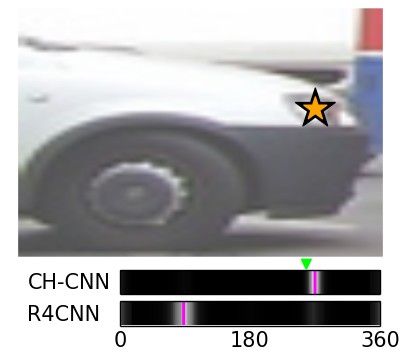}
		\includegraphics[width=.24\linewidth]{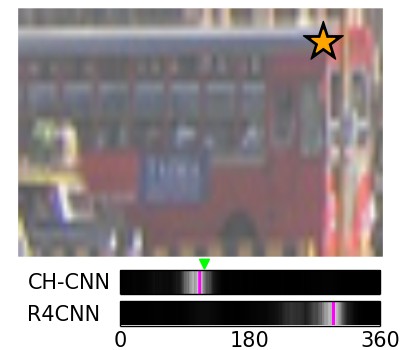}
		\caption{Truncation}
		\label{fig:challenging_examples_truncation}
	\end{subfigure}
	\begin{subfigure}[t]{\linewidth}
		\includegraphics[width=.24\linewidth]{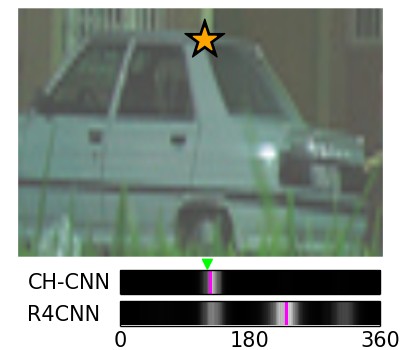}
		\includegraphics[width=.24\linewidth]{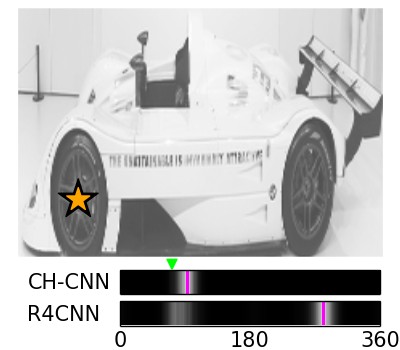}
		\includegraphics[width=.24\linewidth]{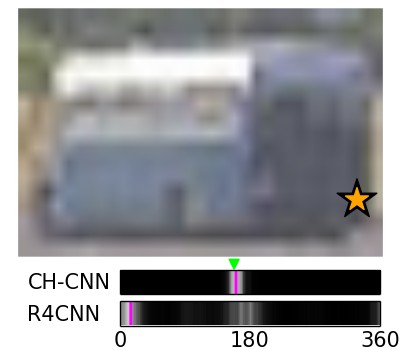}
		\includegraphics[width=.24\linewidth]{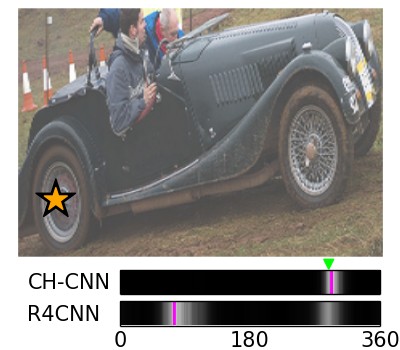}
		\caption{Symmetry}
		\label{fig:challenging_examples_symmetry}
	\end{subfigure}
	\caption{Qualitative comparison of R4CNN~\cite{su_render_2015} and CH-CNN on additional challenging instances.}
	\label{fig:challenging_examples}
\end{figure}

\begin{figure}[t]
	\centering
	\begin{subfigure}[t]{.75\linewidth}
		\includegraphics[width=0.47\linewidth]{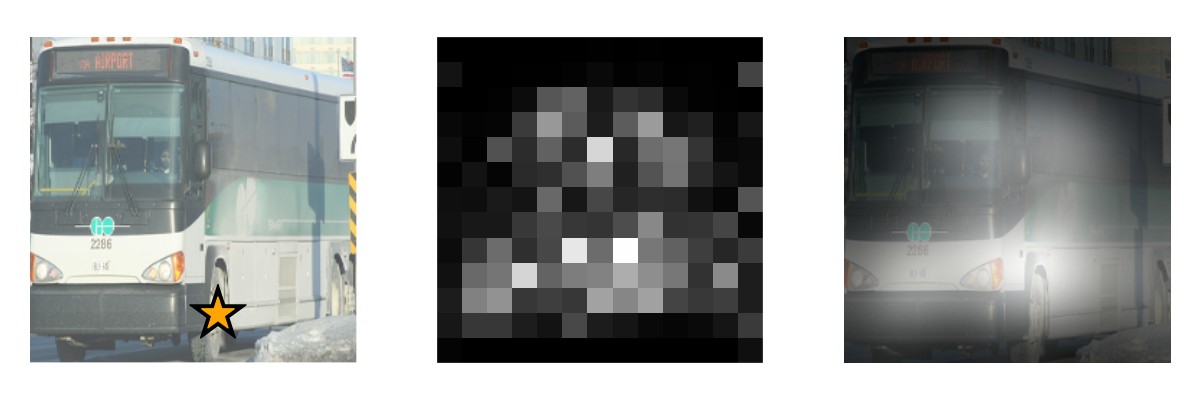}
		\hfill
		\includegraphics[width=0.47\linewidth]{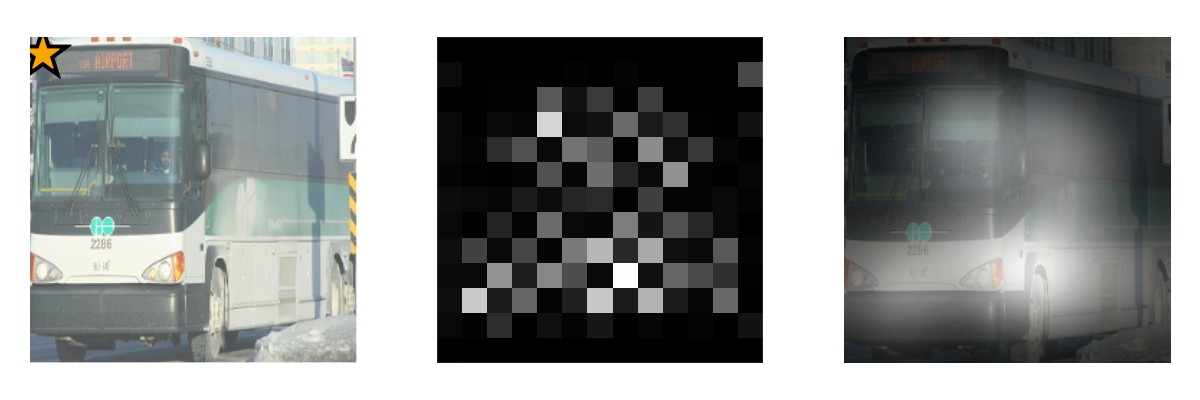}
		\hfill
		\includegraphics[width=0.47\linewidth]{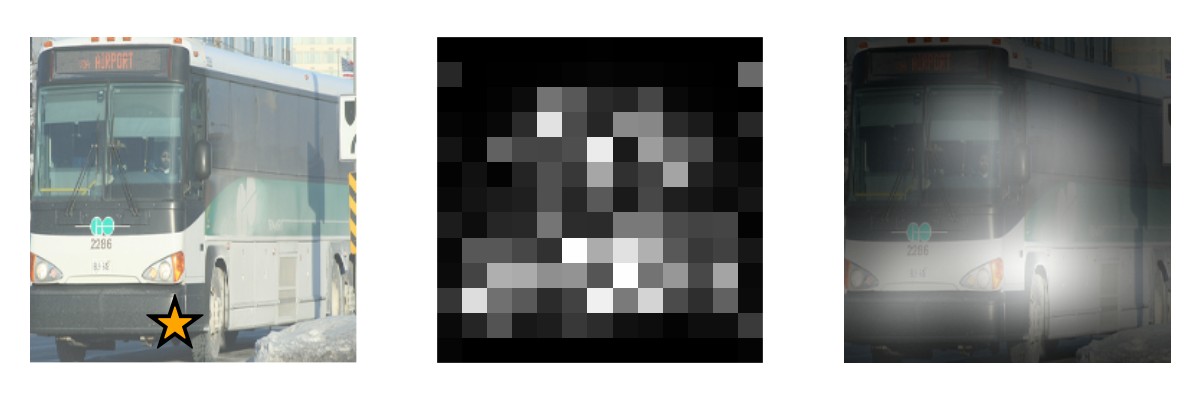}
		\hfill
		\includegraphics[width=0.47\linewidth]{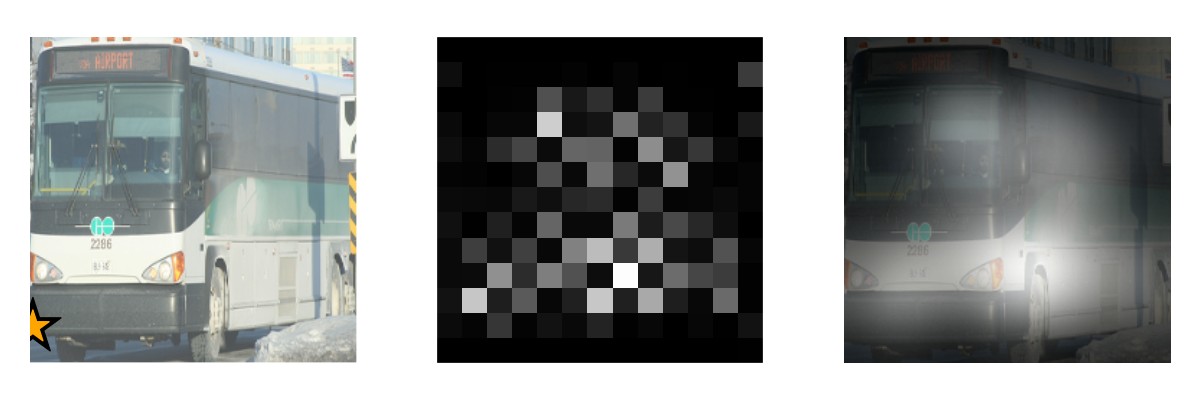}
		\hfill
		\includegraphics[width=0.47\linewidth]{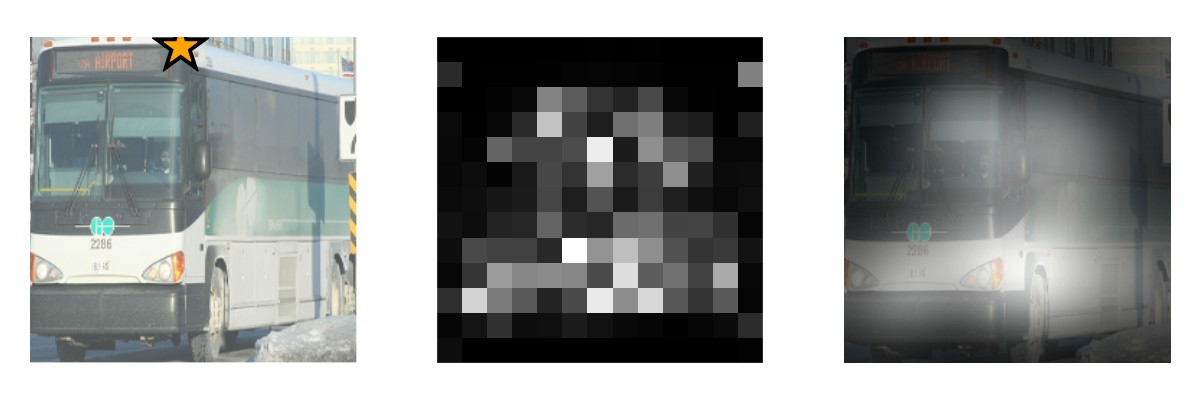}
		\hfill
		\includegraphics[width=0.47\linewidth]{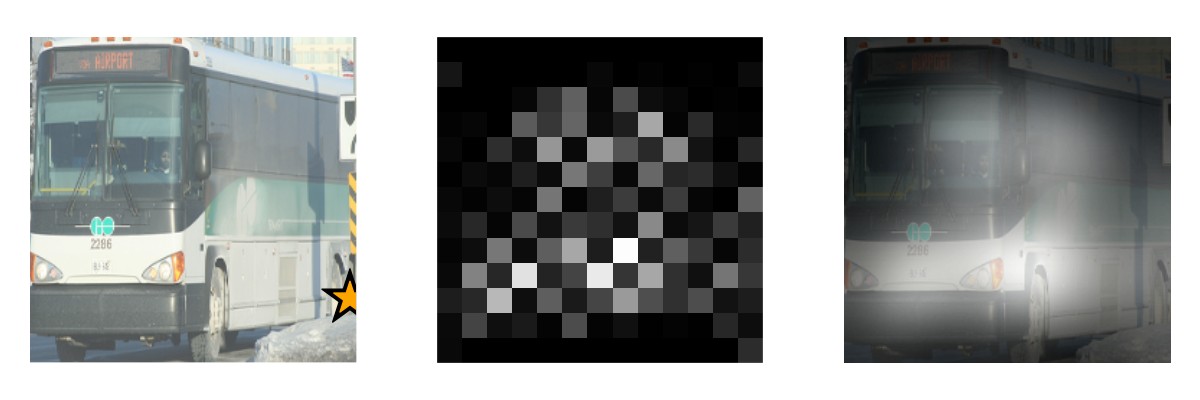}
		\caption{Bus}
	\end{subfigure}
	\begin{subfigure}[t]{.75\linewidth}
		\includegraphics[width=0.47\linewidth]{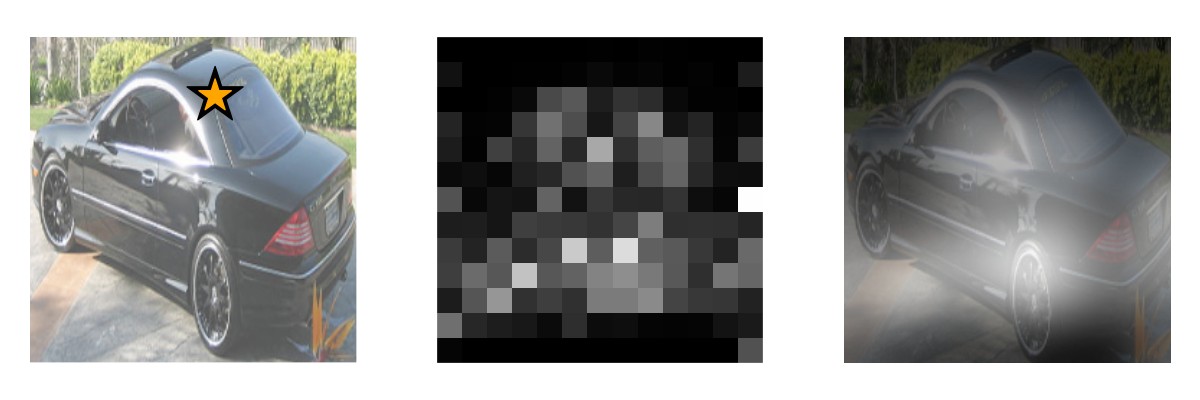}
		\hfill
		\includegraphics[width=0.47\linewidth]{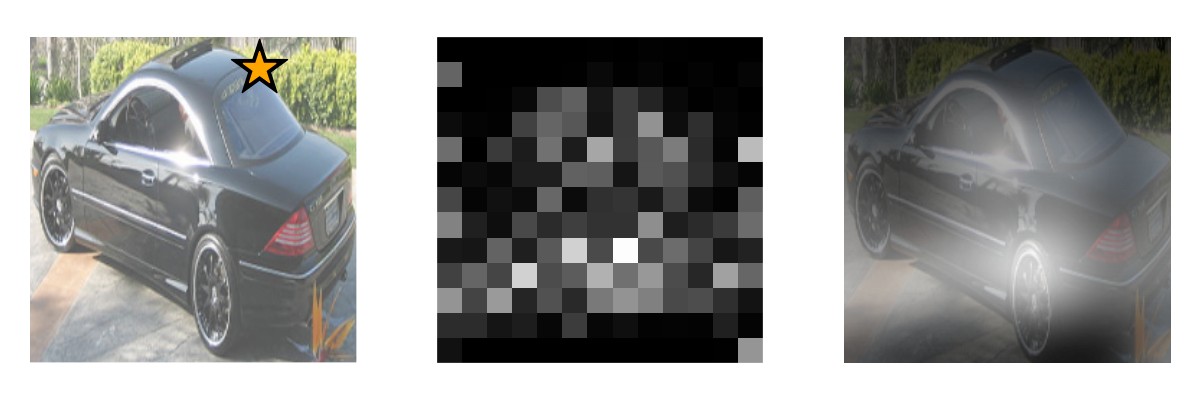}
		\hfill
		\includegraphics[width=0.47\linewidth]{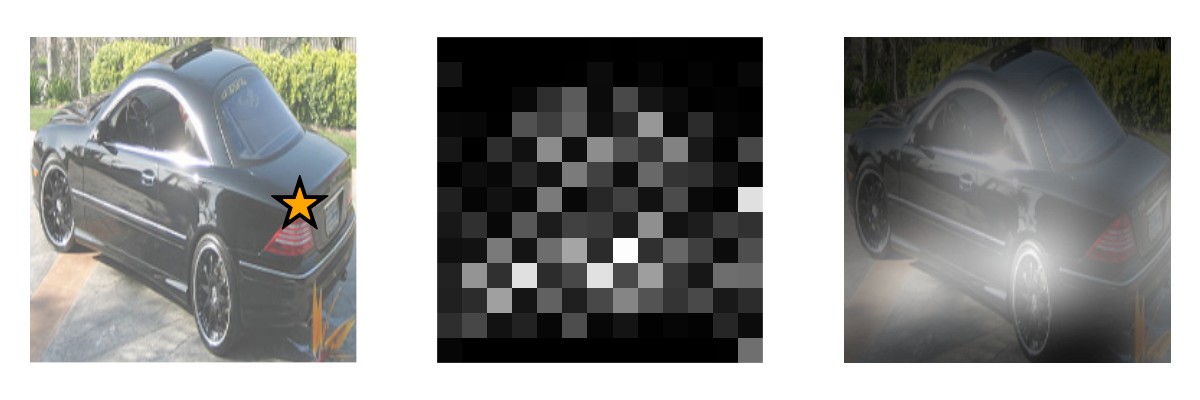}
		\hfill
		\includegraphics[width=0.47\linewidth]{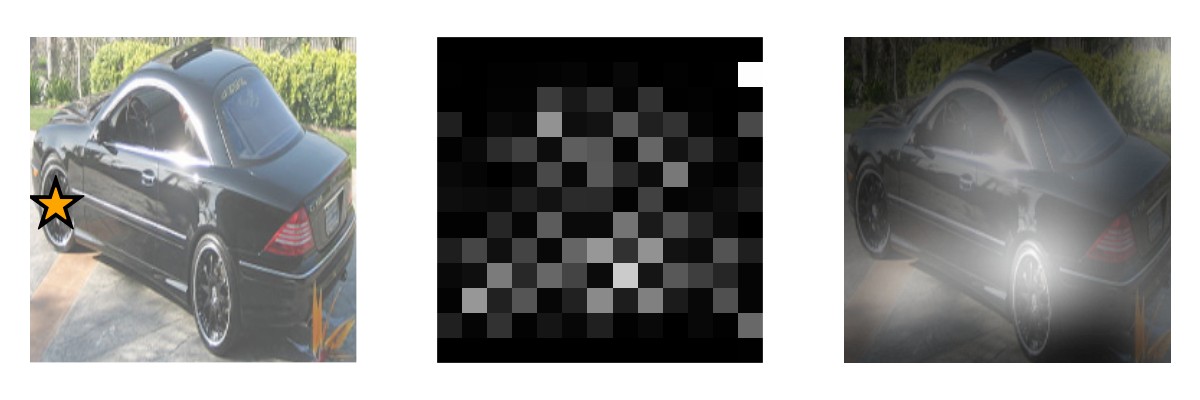}
		\hfill
		\includegraphics[width=0.47\linewidth]{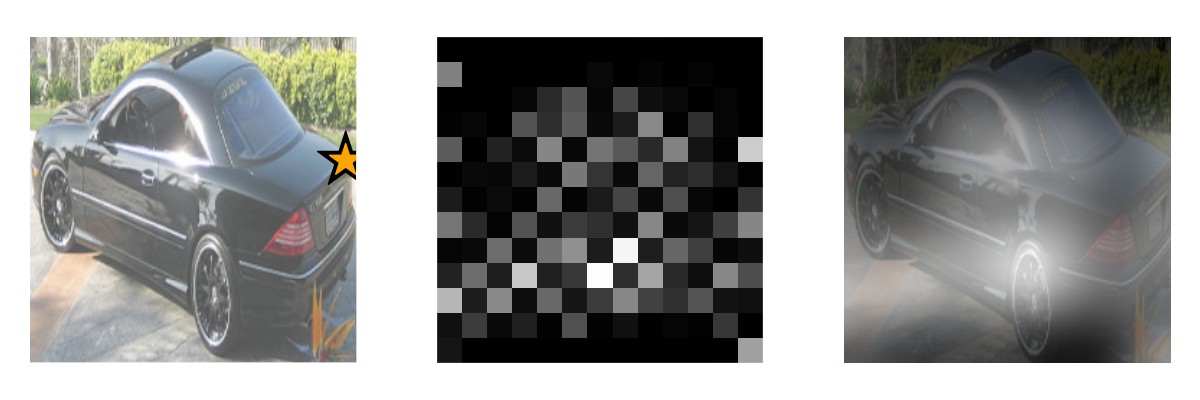}
		\hfill
		\includegraphics[width=0.47\linewidth]{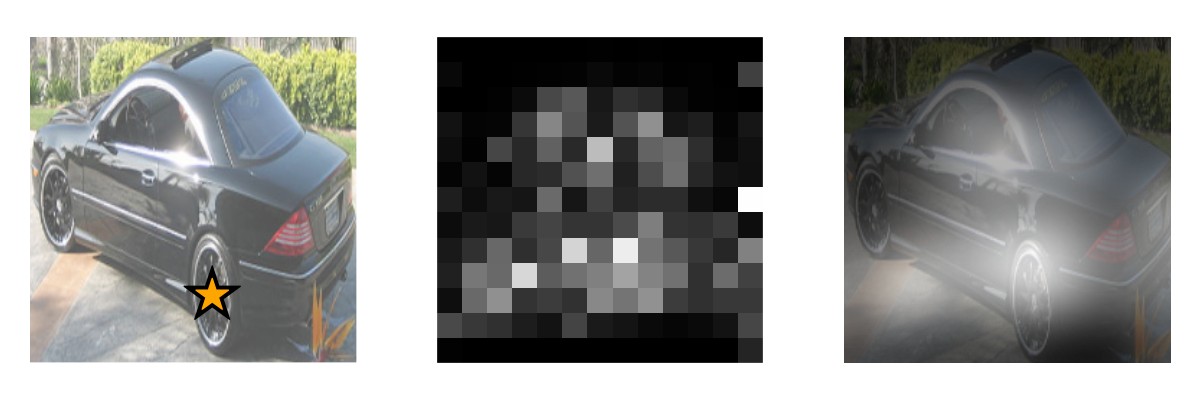}
		\caption{Car}
	\end{subfigure}
	\begin{subfigure}[t]{.75\linewidth}
		\includegraphics[width=0.47\linewidth]{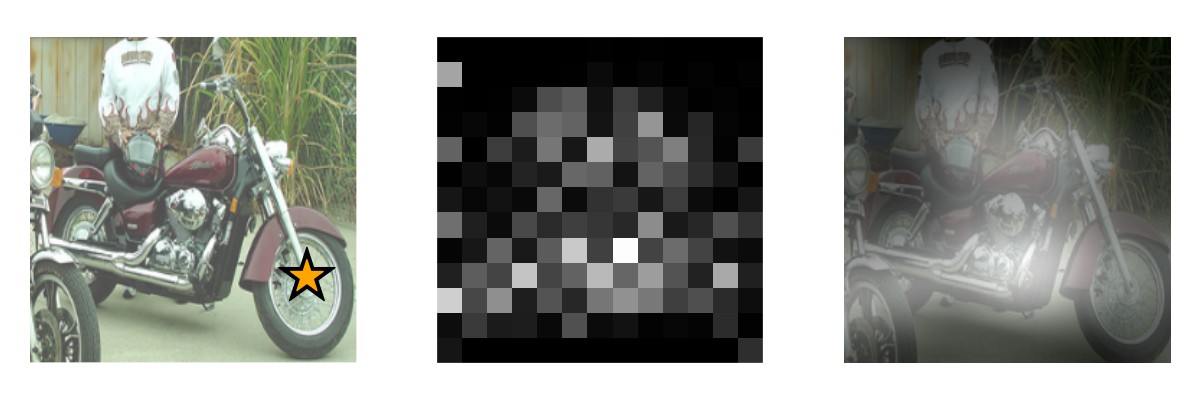}
		\hfill
		\includegraphics[width=0.47\linewidth]{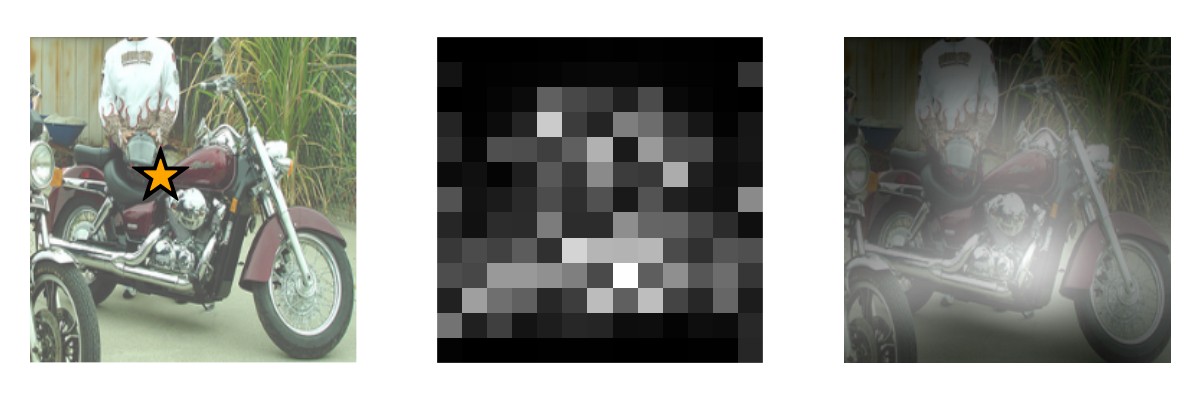}
		\hfill
		\includegraphics[width=0.47\linewidth]{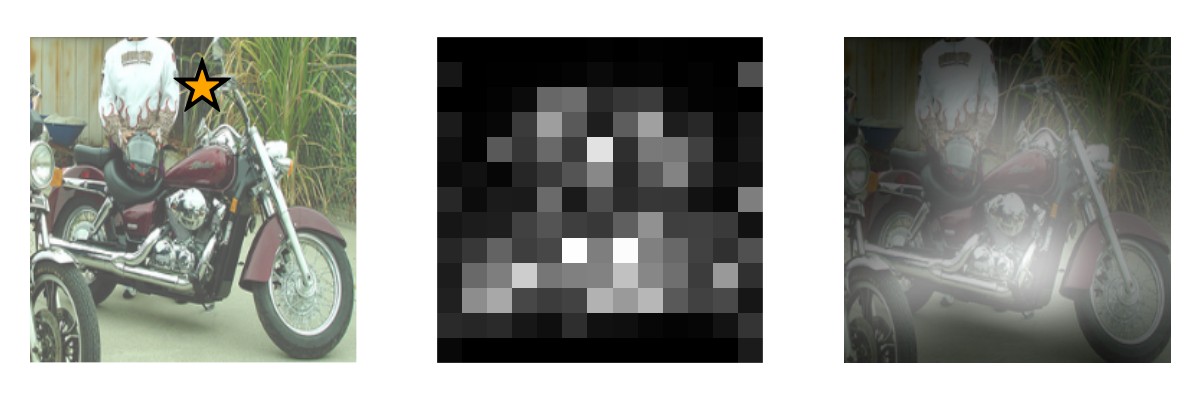}
		\hfill
		\includegraphics[width=0.47\linewidth]{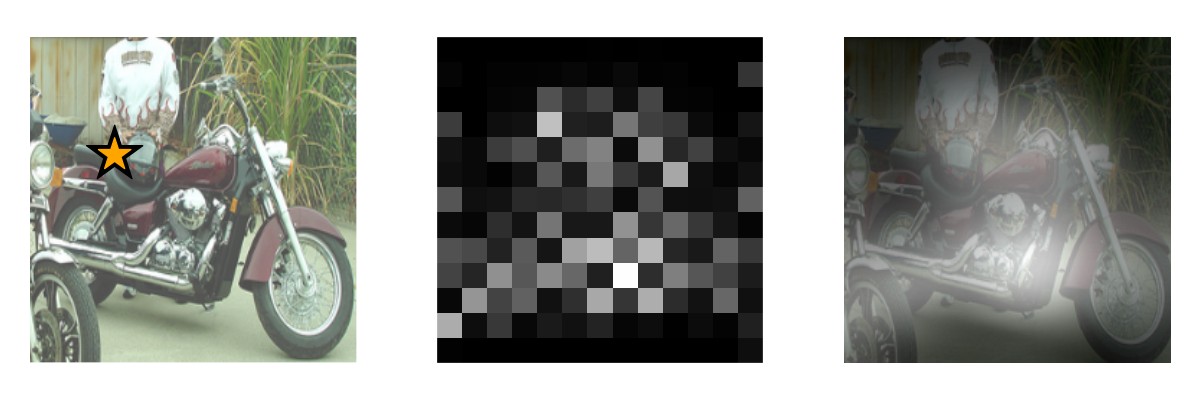}
	\end{subfigure}
	\begin{subfigure}[t]{.8\linewidth}
		\centering
		\includegraphics[width=0.47\linewidth]{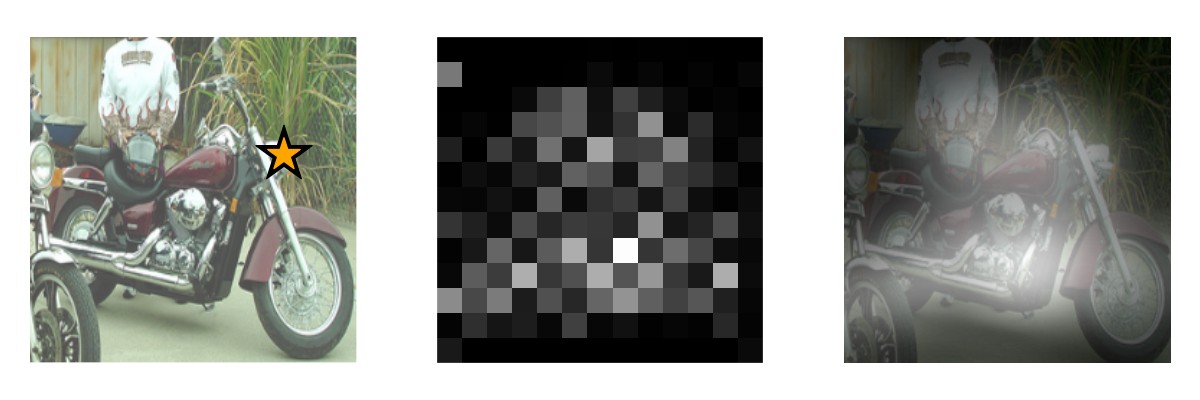}
		\caption{Motorcycle}
	\end{subfigure}
	\caption{The weight maps visualized on test instances from each object class. In each cell, the real image and keypoint visualization is on the left, the learned weight map is in the middle, and a visualization of the weight map overlaid on the image is on the right. The overlay is generated by upsampling the weight map to the original image resolution and applying a Gaussian filter to the result. The light masks and orange stars are for visualizing the keypoint location in this figure only, and are not part of the input to any network.}
	\label{fig:weight_maps}
\end{figure}

In this section, we present additional qualitative results. Figure \ref{fig:challenging_examples} visualizes additional instances where a high degree of occlusion, truncation, or object symmetry is present. Figure \ref{fig:weight_maps} shows the attention maps $\convweightmap$ that are generated from a test instance from each object class.

\newpage

\begin{table}
	\begin{subtable}{\linewidth}
		\centering
		\resizebox{0.75\linewidth}{!}{
			\begin{tabular}{c | c | c | c || c | c | c }
				\multicolumn{1}{c}{} & \multicolumn{3}{c}{$\acc$} & \multicolumn{3}{c}{$MedErr$} \\
				\hline
				Keypoint & R4CNN f.t. & CH-CNN & $\%\uparrow$ & R4CNN f.t. & CH-CNN & $\%\uparrow$ \\
				\hline
				Back left lower corner		& 86.0 & 94.6 & 10.0 & 2.93 & \textbf{2.34} & 20.1 \\
				Back left upper corner		& \textit{81.0} & \textit{91.4} & 12.8 & \textit{3.87} & 2.99 & \textbf{22.7} \\
				Back right lower corner		& 86.7 & 93.9 & 8.30 & 3.32 & 2.92 & 12.1 \\
				Back right upper corner		& 80.9 & 93.0 & \textbf{15.0} & 3.66 & \textit{3.01} & 17.8 \\
				Front left lower corner		& \textbf{94.8} & 97.4 & 2.74 & \textbf{2.64} & 2.55 & 3.41 \\
				Front left upper corner		& 93.3 & 97.1 & 4.07 & 2.66 & 2.61 & \textit{1.88} \\
				Front right lower corner	& 94.4 & 98.0 & 3.81 & 2.83 & 2.55 & 9.89 \\
				Front right upper corner	& 88.2 & 98.6 & 11.8 & 2.83 & 2.66 & 6.01 \\
				Left back wheel				& 93.1 & 97.1 & 4.3 & 2.80 & 2.45 & 12.5 \\
				Left front wheel			& 90.6 & \textbf{100.0} & 10.4 & 3.12 & 2.92 & 6.41 \\
				Right back wheel			& 96.2 & 98.7 & \textit{2.60} & 3.09 & 2.66 & 13.9 \\
				Right front wheel			& 91.5 & \textbf{100.0} & 9.29 & 3.08 & 2.99 & 2.92 \\
				\hline
				\textit{Overall}			& 90.7 & 96.8 & 6.73 & 2.93 & 2.64 & 9.90 \\
				\hline
			\end{tabular}
		}
		\caption{Bus}
		\vspace{8pt}
	\end{subtable}
	
	\begin{subtable}{\linewidth}
		\centering
		\resizebox{0.75\linewidth}{!}{
			\begin{tabular}{c | c | c | c || c | c | c }
				\multicolumn{1}{c}{} & \multicolumn{3}{c}{$\acc$} & \multicolumn{3}{c}{$MedErr$} \\
				\hline
				Keypoint & R4CNN f.t. & CH-CNN & $\%\uparrow$ & R4CNN f.t. & CH-CNN & $\%\uparrow$ \\
				\hline
				Left front wheel		& 86.9 & 89.5 & 2.99 & 5.63 & 5.33 & 5.33 \\
				Left back wheel			& 80.6 & 89.0 & 10.4 & 5.81 & \textit{5.70} & 1.89 \\
				Right front wheel		& 89.4 & 91.2 & 2.01 & 5.73 & \textit{5.70} & \textit{0.52} \\
				Right back wheel		& 85.9 & 90.8 & 5.70 & 5.84 & \textit{5.70} & 2.40 \\
				Left front light		& 90.5 & 94.5 & 4.42 & 4.53 & \textbf{4.24} & 6.40 \\
				Right front light		& \textbf{93.2} & \textbf{95.5} & \textit{2.47} & \textbf{4.50} & 4.27 & 5.11 \\
				Left front windshield	& 87.3 & 91.0 & 4.24 & 5.39 & 4.78 & 11.3 \\
				Right front windshield	& 88.9 & 91.7 & 3.15 & 5.00 & 4.62 & 7.60 \\
				Left back trunk			& 76.8 & 89.5 & 16.5 & 5.87 & 4.70 & 19.9 \\
				Right back trunk		& 72.8 & 88.0 & 20.9 & 5.91 & 4.99 & 15.6 \\
				Left back windshield	& 72.1 & \textit{84.7} & 17.5 & 7.22 & 5.18 & 28.3 \\
				Right back windshield	& \textit{70.8} & 87.6 & \textbf{23.7} & \textit{7.61} & 5.34 & \textbf{29.8} \\
				\hline
				\textit{Overall}		& 78.5 & 90.2 & 14.9 & 5.63 & 4.98 & 11.6 \\
				\hline
			\end{tabular}
		}
		\caption{Car}
		\vspace{8pt}
	\end{subtable}
	
	\begin{subtable}{\linewidth}
		\centering
		\resizebox{0.75\linewidth}{!}{
			\begin{tabular}{c | c | c | c || c | c | c }
				\multicolumn{1}{c}{} & \multicolumn{3}{c}{$\acc$} & \multicolumn{3}{c}{$MedErr$} \\
				\hline
				Keypoint & R4CNN f.t. & CH-CNN & $\%\uparrow$ & R4CNN f.t. & CH-CNN & $\%\uparrow$ \\
				\hline
				Seat back				& 84.2 & 84.7 & 0.59 & 12.6 & 12.5 & 0.79 \\
				Seat front				& 84.3 & 82.6 & \textit{-2.02} & 12.4 & \textit{13.7} & \textit{-10.5} \\
				Head center				& 80.8 & 81.3 & 0.62 & 13.1 & 12.7 & 3.05 \\
				Headlight center		& 90.5 & 90.5 & 0.00 & 10.8 & 10.4 & 3.70 \\
				Back wheel, left side	& 77.5 & 86.5 & 11.6 & 12.3 & 12.3 & 0.00 \\
				Front wheel, left side	& \textbf{91.1} & \textbf{91.9} & 0.88 & \textbf{9.46} & \textbf{9.30} & 1.69 \\
				Left handle end			& 81.8 & \textit{80.3} & -1.83 & 11.5 & 11.7 & -1.74 \\
				Back wheel, right side	& \textit{71.3} & 80.2 & \textbf{12.5} & \textit{14.7} & 12.6 & \textbf{14.3} \\
				Front wheel, right side	& 87.4 & 91.6 & 4.81 & 11.5 & 10.2 & 11.3 \\
				Right handle end		& 84.6 & 83.7 & -1.06 & 11.5 & 11.2 & 2.61 \\
				\hline
				\textit{Overall}		& 84.1 & 85.2 & 1.40 & 11.7 & 11.4 & 2.56 \\
				\hline
			\end{tabular}
		}
		\caption{Motorcycle}
	\end{subtable}
	\caption{Comparison of fine-tuned R4CNN~\cite{su_render_2015} and CH-CNN performance stratified by keypoint class. The $\%\uparrow$ column lists relative increase in performance of CH-CNN over R4CNN. The worst value in each column (lowest for $\acc$ and $\%\uparrow$, highest for $MedErr$) is italicized, and the best value (highest for $\acc$ and $\%\uparrow$, lowest for $MedErr$) is bolded.}
	\label{tab:stratified_performance}
	\vspace{-10pt}
\end{table}

\newpage

\begin{table}
	\begin{subtable}{.32\linewidth}
		\centering
		\resizebox{\linewidth}{!}{
			\begin{tabular}{c | c}
				\hline
				Keypoint & Mean difference \\
				\hline
				Back left lower corner		& 0.1590 \\
				Back left upper corner		& 0.1987 \\
				Back right lower corner		& 0.1418 \\
				Back right upper corner		& 0.2324 \\
				Front left lower corner		& 0.0478 \\
				Front left upper corner		& 0.0713 \\
				Front right lower corner	& 0.0675 \\
				Front right upper corner	& 0.0966 \\
				Left back wheel				& 0.1401 \\
				Left front wheel			& 0.1871 \\
				Right back wheel			& 0.0540 \\
				Right front wheel			& 0.1665 \\
				\hline
				\textit{Overall}			& 0.1140 \\
				\hline
			\end{tabular}
		}
		\caption{Bus}
	\end{subtable}
	\hfill
	\begin{subtable}{.32\linewidth}
		\centering
		\resizebox{\linewidth}{!}{
			\begin{tabular}{c | c}
				\hline
				Keypoint & Mean difference \\
				\hline
				Left front wheel		& 0.0701 \\
				Left back wheel			& 0.1389 \\
				Right front wheel		& 0.0306 \\
				Right back wheel		& 0.0777 \\
				Left front light		& 0.0862 \\
				Right front light		& 0.0625 \\
				Left front windshield	& 0.0715 \\
				Right front windshield	& 0.0467 \\
				Left back trunk			& 0.2601 \\
				Right back trunk		& 0.2949 \\
				Left back windshield	& 0.2681 \\
				Right back windshield	& 0.3110 \\
				\hline
				\textit{Overall}		& 0.1532 \\
				\hline
			\end{tabular}
		}
		\caption{Car}
	\end{subtable}
	\hfill
	\begin{subtable}{.32\linewidth}
		\centering
		\resizebox{\linewidth}{!}{
			\begin{tabular}{c | c}
				\hline
				Keypoint & Mean difference \\
				\hline
				Seat back				&  0.0062 \\
				Seat front				& -0.0091 \\
				Head center				&  0.0151 \\
				Headlight center		&  0.0223 \\
				Back wheel, left side	&  0.1653 \\
				Front wheel, left side	&  0.0187 \\
				Left handle end			& -0.0353 \\
				Back wheel, right side	&  0.1739 \\
				Front wheel, right side	&  0.0660 \\
				Right handle end		& -0.0268 \\
				\hline
				\textit{Overall}		& 0.0240 \\
				\hline
			\end{tabular}
		}
		\caption{Motorcycle}
	\end{subtable}
	\caption{Mean difference in absolute error for individual instances (R4CNN - CH-CNN) stratified by keypoint class.}
	\label{tab:stratified_error}
\end{table}

\newpage

\newpage

\newpage

\end{appendices}

\end{document}